\documentclass[10pt,twocolumn,letterpaper]{article}

\usepackage{cvpr}              % To produce the CAMERA-READY version
\usepackage{diagbox}
\usepackage{graphicx}
\usepackage{amsmath}
\usepackage{amssymb}
\usepackage{booktabs}
\usepackage[dvipsnames]{xcolor}
\usepackage{flexisym}
\usepackage{amsmath}
\usepackage{multirow}
\usepackage{tabularx}
\usepackage{enumitem}

\usepackage{amssymb}% http://ctan.org/pkg/amssymb
\usepackage{pifont}% http://ctan.org/pkg/pifont
\usepackage[dvipsnames]{xcolor}

\definecolor{cvprblue}{rgb}{0.21,0.49,0.74}
\usepackage[pagebackref,breaklinks,colorlinks,citecolor=cvprblue]{hyperref}

\def\paperID{3} % *** Enter the Paper ID here
\def\confName{CVPR}
\def\confYear{2024}

\title{MonoSelfRecon: Purely Self-Supervised Explicit Generalizable 3D Reconstruction of Indoor Scenes from Monocular RGB Views}

\author{Runfa Li\textsuperscript{1} \quad
Upal Mahbub\textsuperscript{2} \quad
Vasudev Bhaskaran\textsuperscript{2} \quad
Truong Nguyen\textsuperscript{1} \quad
\\
\textsuperscript{1}UC San Diego \quad
\textsuperscript{2}Qualcomm %Technologies Inc.
}

\begin{document}
\maketitle
\begin{abstract}
 Current monocular 3D scene reconstruction (3DR) works are either fully-supervised, or not generalizable, or implicit in 3D representation. We propose a novel framework - \textbf{MonoSelfRecon} that for the first time achieves \textbf{explicit 3D mesh} reconstruction for \textbf{generalizable} indoor scenes with monocular RGB views by purely \textbf{self-supervision} on voxel-SDF (signed distance function). MonoSelfRecon follows an Autoencoder-based architecture, decodes voxel-SDF and a generalizable Neural Radiance Field (NeRF), which is used to guide voxel-SDF in self-supervision. We propose novel self-supervised losses, which not only support pure self-supervision, but can be used together with supervised signals to further boost supervised training. Our experiments show that "MonoSelfRecon" trained in pure self-supervision outperforms current best self-supervised indoor depth estimation models and is comparable to 3DR models trained in fully supervision with depth annotations. MonoSelfRecon is not restricted by specific model design, which can be used to any models with voxel-SDF for purely self-supervised manner. %\footnote{Datasets  were  downloaded  and  evaluated  solely  by  the researchers from UC San Diego.}
\end{abstract}    
\vspace{-5mm}
\section{Introduction}
\label{sec:intro}

\begin{figure*}[t]
    \centering
    \includegraphics[width=\linewidth]{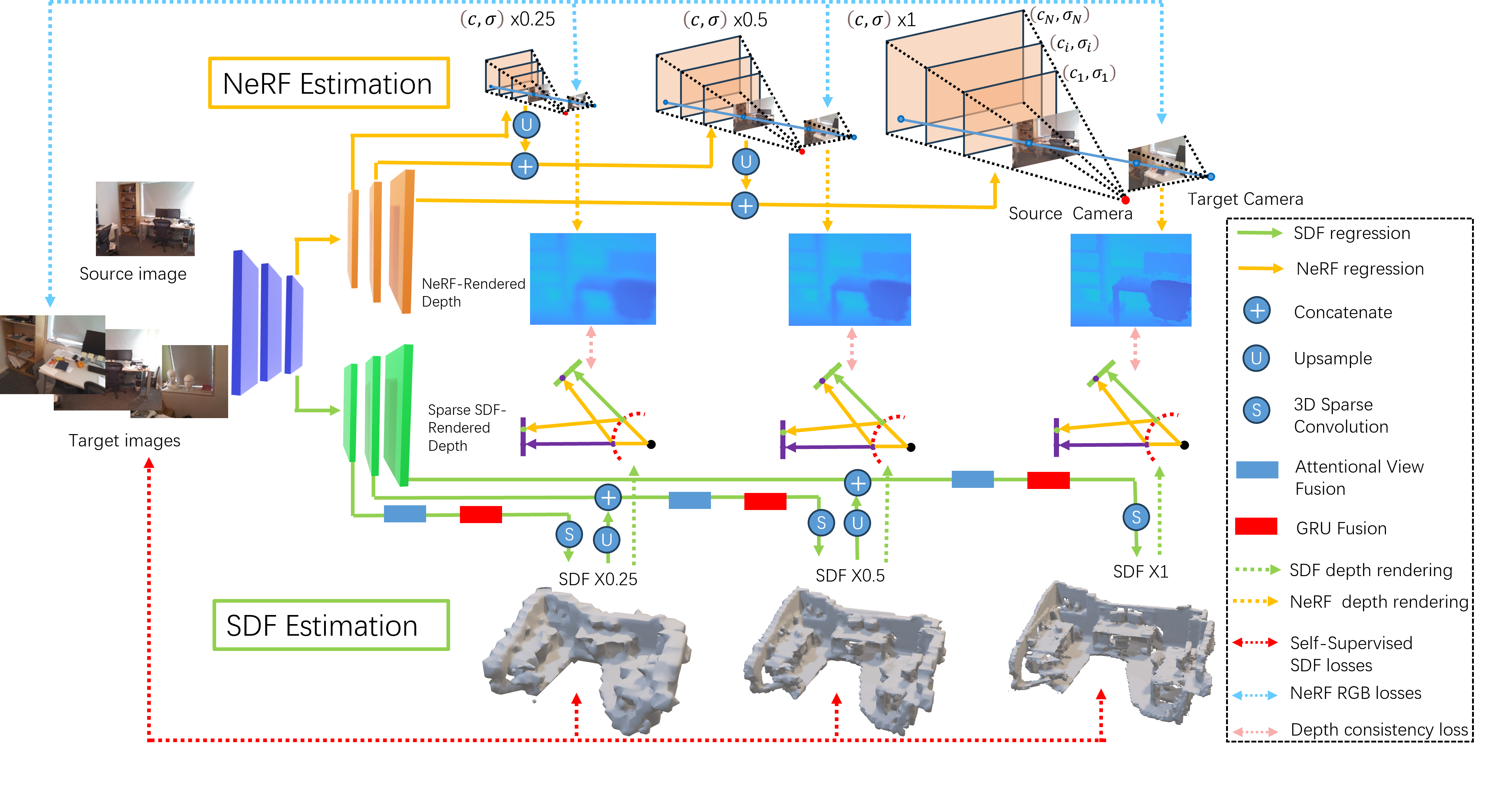}
    \vspace{-12mm}
    \caption{\textbf{MonoSelfRecon Pipeline.} We use monocular RGB sequence to estimate explicit 3D mesh of generalizble indoor scenes under purely self-supervision free of SDF or depth annotations. A coarse-to-fine architecture is used for both SDF and NeRF decoders, where our proposed self-supervised losses are used between SDF-inputRGB, NeRF-inputRGB, and SDF-NeRF.}
    \label{fig:pipeline}
    \vspace{-5mm}
\end{figure*}

\quad 3D scene reconstruction is one of the most important problems in 3D computer vision, robotics, virtual reality, autonomous driving and so on. Indoor 3DR is even more challenging because of the large texture-less region and complicated relations between objects. As more complicated and accurate models emerge, the main concern shifts to three standards: 1. Self-supervised training without laborious large-amount of annotations in depth or SDF, 2. Generalizable to same type of scenes, and 3. Explicit 3D mesh representations.

\begin{table}[]
\setlength\tabcolsep{1pt}
\footnotesize
\begin{tabular}{l|c|c|c}
\hline
\diagbox{Methods}{Standards}             & \multicolumn{1}{l|}{Self-Supervised} & \multicolumn{1}{l|}{Generalizable} & \multicolumn{1}{l}{Explicit 3D Mesh} \\ \hline
Self-supervised Depth & \textbf{\checkmark} & \textbf{\checkmark} & \ding{55} \\
Supervised Depth      & \ding{55} & \textbf{\checkmark}  & \ding{55}    \\
Supervised Voxel-SDF  & \ding{55} & \textbf{\checkmark}  & \textbf{\checkmark}    \\
Implicit NeRF         & \textbf{\checkmark}  & \ding{55} & \ding{55}          \\
Explicit SDF-NeRF     & \textbf{\checkmark}  & \ding{55} & \textbf{\checkmark}   \\ \hline
\textbf{Ours}  & \textbf{\checkmark}  & \textbf{\checkmark} & \textbf{\checkmark}  \\ \hline
\end{tabular}
\vspace{-3mm}
\caption{\textbf{Monocular 3DR measured in three major standards.}}
\label{table:3dr class}
\vspace{-7mm}
\end{table}

Explicit monocular 3DR models originate from depth estimation \cite{kineticfusion, 3dgo, vif, mvdepthnet, smartphone, neuralrgb, mvstempo}, where depth maps can be directly supervised with pixel-wise depth ground truth. Depth maps can be also self-supervised by continuous RGB sequences from multiple camera views, with or without camera poses. Under constraints such as temporal smoothness, multiple key frames are selected from continuous depth map sequence, fused to global 3D volume and integrate ``Truncated Signed Distance Function" value (TSDF) \cite{kineticfusion} for each voxel within the truncation distance. 3D surface mesh can be obtained from ``marching cube'' algorithm with estimated TSDF \cite{marchingcube}. However, because of the depth inconsistency in multi-views, directly fusing depth estimation for TSDF may cause it either layered or too sparse. Additionally, using depth maps from multiple views for TSDF fusion typically causes large overlapping regions, which is a waste of computation in inference\cite{neucon}. Later works \cite{atlas, neucon, vortx} overcome the limitations by pre-defining 3D voxels and directly regressing SDF for each voxel. However, these works must be trained in supervision with laboriously large amount of fine-grained SDF annotations fusing from depth map ground truth.

Implicit monocular 3DR works start to dominate with the emergence of NeRF\cite{nerf}. One advantage of implicit 3DR over explicit ones is its purely self-supervised training. With RGB sequence from different camera poses, NeRF reconstructs scenes in implicit representation by synthesizing images from new views. Recently, State-of-the-art (SOTA) works introduce implicit SDF function to NeRF, where the SDF-NeRF can estimate explicit SDF values w.r.t any specified 3D position in the scene \cite{hnerf, isdf, manhattansdf, monosdf, volsdf, neuris}. However, these 3DR works are still non-generalizable to other scenes. Although few works such as \cite{pixelnerf,mpi-nerf} make NeRF generalizable, the generalization has not been extended to estimate explicit SDF values.

We summarize current 3DR methods based on the three standards in Table \ref{table:3dr class}, where each type of the method is discussed in section \ref{sec:related_works}. Different to all existing works, where the 3DR either requires large amount of annotated ground truth, or non-generalizable, or not explicit in 3D representation, our work overcomes these three concerns altogether. Our key contributions are:

\begin{itemize}[leftmargin=0.2cm]
\setlength{\itemsep}{0pt}
\setlength{\parsep}{0pt}
\setlength{\parskip}{0pt}

\item We propose a novel framework ``MonoSelfRecon'' that first time achieves \textbf{explicit 3D mesh} reconstruction for
\textbf{generalizable} indoor scenes with monocular RGB sequence by purely \textbf{self-supervised} training on voxel-SDF, which is different to all existing works.

\item We propose novel self-supervised losses and conduct purely self-supervised training without SDF or depth ground truth. Experiments show that our approach outperforms the best self-supervised generalizable indoor 3DR (depth-based) works, and is comparable to fully-supervised works.

\item Our proposed self-supervised losses can be used together with supervised losses to boost fully supervised training. The framework is not limited to specific models but free to extend to any models with voxel-SDF estimation. Consequently, it keeps the advantages of the original model, such as inference speed and meomory efficiency.

\end{itemize}

\section{Related Works}
\label{sec:related_works}

\iffalse
To show the difference and improvement of our work to other 3DR works, we measure 3DR into three major standards based on their training strategies(supervised/unsupervised), generalization abilities, and representations(explicit/implicit), as shown in Table \ref{table:3dr class}.
\fi

In this section, we discuss details of each method based on the three standards in Table \ref{table:3dr class}. 

\noindent
\textbf{Generalizable Explicit depth Reconstruction}.
Initially, 3D mesh reconstruction relies on depth estimation, the typical way is to perform TSDF fusion from depth and use marching cube to recover mesh from TSDF. Fully supervised training is the first choice when available depth map ground truth is used. These models take as input single image, and explore different model designs \cite{deepv2d,CNMNet,neuralrgb,mvdepthnet,dpsnet}, but all regress to 2D depth map and directly supervise with depth ground truth. However, since  annotations are missing and unreliable in many cases, self-supervision has to be used without depth annotations. With camera poses, the depth can be estimated by binocular stereo matching, leveraging epipolar-geometric consistent matching between image pixels \cite{mvdepthnet}, or replacing pixels with extracted deep features to produce a matching cost volume \cite{ESTDepth, deepvideomvs}. Without camera poses, it becomes a structure-from-motion (SFM) problem, the network can jointly estimate depth and pose with reconstruction error \cite{monoindoor,p2net,movingindoor,structdepth,monodepth2}. However, self-supervised depth has scale drifting and ambiguity problems, which need other priors to recover real scale. Alternatively, we directly regress voxel-SDF from network in real scale.

\noindent
\textbf{Generalizable Supervised Explicit 3D mesh Reconstruction}.
Directly regressing voxel-SDF is generalizable and straightforward, the advantages are real-scale, time and memory efficient. However the huge volume of 3D convolution \cite{atlas} requires strong supervise signals - voxel-SDF ground truth fused from depth ground truth, although later works \cite{neucon, vortx} shrink the volume by using 3D sparse convolution, SDF ground truth is still inevitable, which is not easy to obtain in many cases. In this work, we explore pure self-supervision to train voxel-SDF free from any SDF or depth annotations.

\noindent
\textbf{Unsupervised implicit 3D Reconstruction}. NeRF \cite{nerf} is the representative of implicit 3DR. Many works \cite{mvsnerf, pointnerf, pixelnerf, nsvf, plenoxels, instant, blocknerf, nerfusion} improve the baseline NeRF from different aspects. MVSNeRF \cite{mvsnerf} speeds up NeRF in a cost volume built within a multi-view-system, while PointNeRF \cite{pointnerf} speeds up by introducing point cloud from estimated depth maps to sample the ray-level key points only with the nearby point cloud. However, all these works require per-scene optimization and are limited to implicit RGB synthesis. PixelNeRF \cite{pixelnerf} improves volume rendering function by allowing rays from one view to refer to key points on rays from other views, which endows NeRF a limited generalization by using fewer views. Later works \cite{mpi-nerf, ibrnerf} explored increasing generalization by adding an explicit image encoder on top of positional encoding that enables NeRF to generalize to similar but unseen scenes. However, these works are still implict NeRF where the outputs are synthesized RGB image, unlike our desired explicit 3D mesh.

\noindent
\textbf{Non-Generalizable Unsupervised Explicit 3D mesh Reconstruction}.
The standard NeRF for implicit 2D view synthesis has been successfully extended to explicit 3D mesh representation \cite{hnerf, isdf, manhattansdf, monosdf, volsdf, neuris, neus, neuralangelo, neuralwarp, unisurf}. With specified 3D coordinates and pose as input, these models estimate SDF values wrt. ray-level key points. The core design is a density function converting per-point SDF estimation to opacity values for volumetric rendering \cite{volsdf}, where the ray-level SDF values are queried to obtain rendered pixel intensity, which brings SDF to the reconstruction loss. Later works add more priors to supervise together with SDF-NeRF loss. \cite{manhattansdf} uses supervised pre-trained depth and semantic segmentation estimator, while \cite{monosdf} and \cite{neuris} use supervised pre-trained depth estimator and surface normal detector. However, the generalization ability shown in implicit NeRF has not been successfully extended to explict SDF-NeRF, which means once trained, the SDF-NeRF can only estimate 3D mesh of a specific scene. Our work aims to keep the explicit representation, self-supervised training, while enables generalization.

\noindent
\textbf{Motivation.} The question is that If these per-scene optimization methods are fast enough, then there's no disadvantage to them. However, non-generalizable methods take a long time to converge per scene. Neuralangelo\cite{neuralangelo} takes over 4 hours(single 3090Ti) to converge on a simple lego toy and almost a day for a big room, which is comparable to the generalized methods (ours 36 hours training on single 3090Ti but once trained can be used on different indoor scenes). Non-generalizable methods can be used where a scene is accessed repeatedly and unchanged, such as a city block, long training is acceptable as once trained it can be used for a while unless the scene changes. However, generalizable methods are better when different scenes need to be reconstructed ASAP and no time for per-scene optimization for hours, such as a real estate agent visits 20 different buildings in a day to scan 3D mesh models for unseen rooms and send it to clients right away. One compromise is part optimization but it is only suitable for unchanged scenes, whenever the scene is changed, they need to retrain the NeRF of corresponding part, while ours don’t need retrain and can infer the new building in real time. We adapted NeuralRecon as backbone, our inference speed is 30 ms per frame (single 2080Ti).

\begin{figure*}
\begin{minipage}{0.55\linewidth}
  \centerline{\includegraphics[width=1.0\textwidth]{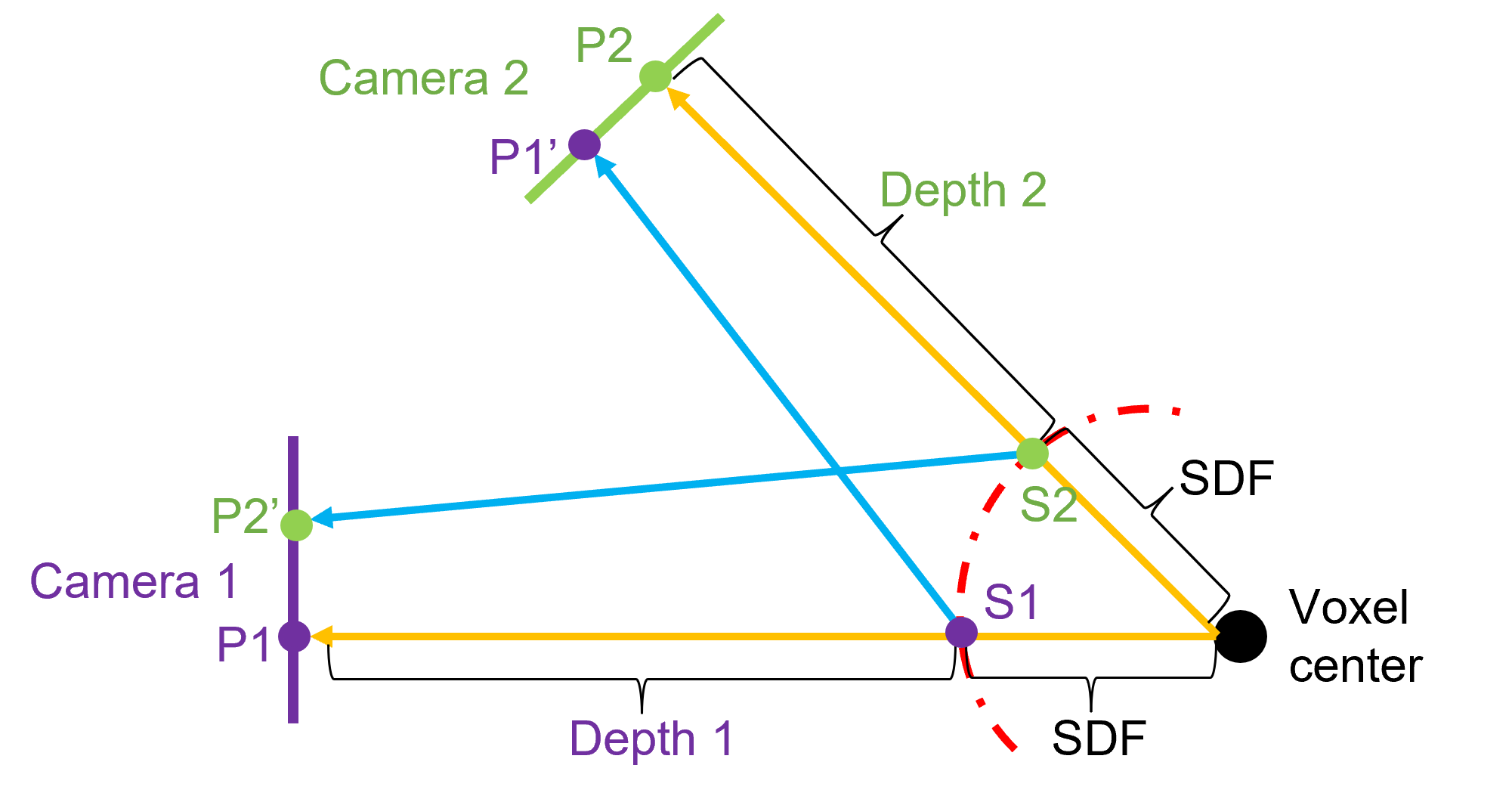}}
\end{minipage}
\hfill
\begin{minipage}{0.51\linewidth}
  \centerline{\includegraphics[width=1.0\textwidth]{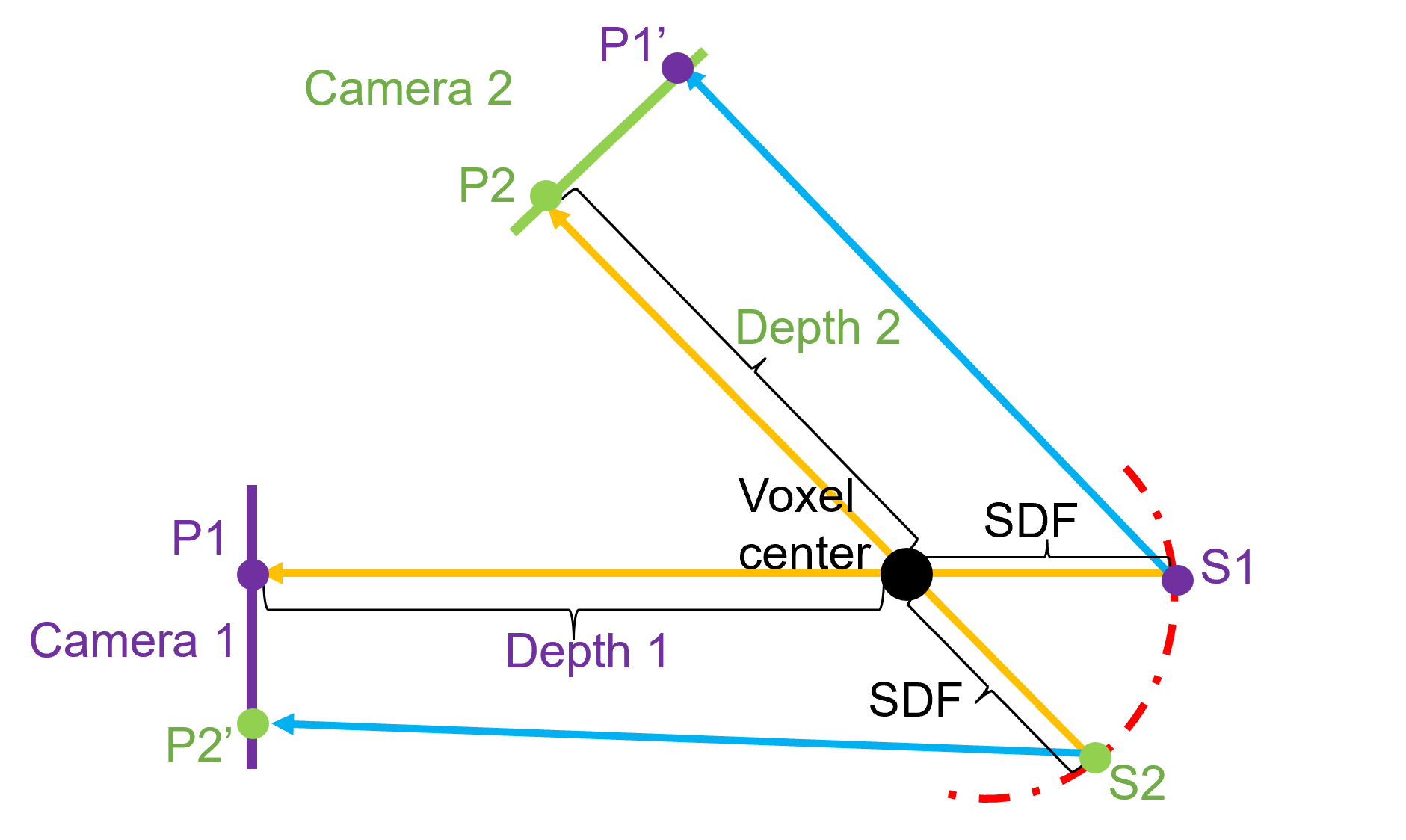}}
\end{minipage}
\vspace{-5mm}
\caption{\textbf{Self-supervised SDF photometric loss between source and target views}. \textbf{Left:} Voxel center is inside of the surface, SDF is negative. Orange arrows show projecting rays from voxel centers to 2D pixels (P1, P2) on each camera plane, blue arrows show reprojection of surface 3D points (S1, S2) to 2D pixels (P1', P2') in each camera plane. Surface points are estimated by SDF estimation. \textbf{Right:} Voxel center is outside of the surface, SDF is positive. \textbf{The loss is extended to all n views in a fragment}.}
\label{fig:sdf photometric loss}
\vspace{-5mm}
\end{figure*}

\vspace{-2mm}
\section{Method}
\label{sec:method}
\vspace{-2mm}
\subsection{MonoSelfRecon Framework}
\quad Figure \ref{fig:pipeline} shows our MonoSelfRecon framework. In both training and testing, we take the input monocular RGB sequence and camera poses, and reconstruct 3D mesh of the whole scene. We select key frames following the process in \cite{key_select}, where we consider a valid frame to be ``greater than 15 degree in rotation or 0.3 meter in translation'' to the previous frame. Every $n$ consecutive key frames form a scene fragment. Fragments are fed as inputs, from which, the network extracts 2D features per key frame, creates 3D feature volume and jointly estimates SDF and NeRF using separate decoders at three pyramid scales. In SDF decoder, 2D features are fused to 3D voxel features and regress to voxel-SDF values at each level with 3D sparse convolution. Every time the network only estimates SDF corresponding to the fragment in a $[N,N,N]$ 3D voxel region, the Gated Recurrent Unit (GRU) module at each level updates SDF, fuses reconstruction from previous fragments, and completes the whole scene. During training, self-supervised losses are implemented between SDF-input, NeRF-input, and SDF-NeRF, with detailed discussion in \ref{sec:losses}. During testing, 3D mesh can be obtained from SDF through marching cube\cite{marchingcube}.  

\noindent
\textbf{Attentional View Fusion}.
For each fragment, 3D voxel features can be simply obtained by projecting the 3D voxel to each 2D view in the fragment, searching for visible corresponding pixels, and averaging 2D features. However, since each view have different distance, angle, and occlusion to voxels, 2D features from different views should not contribute the same to 3D features. Inspired by recent works \cite{vortx}, we use an attentional view fusion module by adding a light transformer before averaging features. The transformer takes unordered sequence of 2D features and updates weighted features before average to 3D, which enables more flexibility to adjust the contribution of each frame to the fragment. Although simple, this module achieves significant improvement as shown in the ablation study in Table \ref{table:scannet_ablation}.

\noindent
\textbf{GRU}. 
We adapt the GRU module from \cite{neucon}. With camera poses, it is simple to concatenate fragments and replace the overlapping voxels of latter fragment to the previous one, but it ignores the effect of the latter views to the previous ones. Once fusing 3D voxel features from 2D and before regressing voxel-SDF at each level, the GRU module takes 3D voxel features of both previous and current fragments as input to update the current features. GRU fusion makes an obvious improvement as shown in the ablation study in Table \ref{table:scannet_ablation}.

\noindent
\textbf{NeRF.} We adopt the generalizable MPI (MultiPlane-Images)-NeRF \cite{mpi, mpi-nerf, mononerf}, introduce it to the framework and jointly train with SDF to boost SDF estimation. The core idea is to use an explicit encoder on top of standard positional encoding to enable implicit NeRF with generalization ability. The NeRF estimation also further boosts our SDF performance as shown in the ablation study in Table \ref{table:scannet_ablation}.

\subsection{Self-supervised Losses}
\label{sec:losses}

\noindent
\textbf{SDF Photometric Loss}. Figure \ref{fig:sdf photometric loss} shows a simplified version of the loss implementation between two camera views, where the corresponding 2D coordinates (P1 and P2) can be found by tracing rays (orange arrows) to camera planes. SDF is the distance between a point to its nearest surface, where the value is negative when the point is inside of the surface, and positive when it is outside of the surface. The model estimates SDF per voxel. The depth $\hat{D}_{cam}$ can be estimated by Eq. \ref{eq:get_depth}, where $V_{world}$ is 3D world coordinate of the pre-defined voxel center, $T_{world\xrightarrow{}cam}$ is camera extrinsic, $\hat{SDF}$ is the estimated voxel-SDF from the model. With depth and voxel center in the camera coordinate, the surface points S1 and S2 can be estimated by Eq. \ref{eq:get_surface}, where $\vec{ray}$ is the unit vector at ray direction (orange arrows), and $\hat{S}_{cam}$ is the 3D coordinate of a surface point in the camera view. Finally, the reconstructed pixels are obtained by reprojecting (blue arrows) surface points S1, S2 to camera 2 and 1 as Eq. \ref{eq:point_proj}, where K is camera intrinsic, $T_{cam\xrightarrow{}cam'}$ is camera pose from cam to cam'. P-P' is a pixel reconstruction pair with same photometric intensity, where a photometric consistency loss can be derived as Eq. \ref{eq:pts_loss}. The exact pixel intensities $I_{cam}(P)$ and $I_{cam'}(P')$ are obtained with bilinear interpolation from projected 2D points lying between integer coordinates, and the loss is only traced to the points lying within the camera planes.
\vspace{-3mm}
\begin{equation}
\begin{split}
    V_{cam} = T_{world\xrightarrow{}cam}V_{world} \\
    \hat{D}_{cam} = V_{cam} + \hat{SDF}
    \label{eq:get_depth}
\end{split}
\end{equation}
\vspace{-6mm}

\vspace{-5mm}
\begin{equation}
\begin{split}
    \hat{S}_{cam}(x, y, z) = V_{cam}(x, y, z) + |\hat{SDF}| \Vec{ray}(x, y, z)
    \label{eq:get_surface}
\end{split}
\end{equation}
\vspace{-6mm}

\vspace{-5mm}
\begin{equation}
\begin{split}
    P = Interp(KV_{cam}) \\
    P' = Interp(K'T_{cam\xrightarrow{}cam'}\hat{S}_{cam})
\end{split}
\label{eq:point_proj}
\end{equation}
\vspace{-6mm}

\vspace{-4mm}
\begin{equation}
    L_{sdf} = \sum_{P \in cam}\sum_{P'\in cam'}{|(I_{cam}(P) - I_{cam'}(P')|}
\label{eq:pts_loss}
\end{equation}
\vspace{-4mm}

\iffalse
We make an assumption to set up this point-wise self-supervised SDF loss. Although the distance from the voxel center to different surface points varies - only the nearest distance is SDF. Since the network estimates one SDF value per voxel, we use the same estimated SDF value corresponding to the same voxel center to estimate surface points for different camera views. However, previous supervised works made the same assumption to implement TSDF fusion  \cite{atlas, neucon} and get TSDF ground truth from the depth map. More specifically, instead of using the nearest distance, they take the average distances from all views. Similarly, we assign this task to the network, when the resolution is large enough (voxel size is small enough), the network is trained to get to the average distance that optimizes the consistency loss between all camera views.
\fi

In practice, we implement the loss across all views in the scene fragment and take the weighted average loss, where the weight is in direct proportion to the number of the candidate P-P' pair. If P' lies outside of the other camera's plane, we ignore this P-P' pair. For SFM-based self-supervised depth works, they start from 2D pixel and ends up at 2D pixel to jointly regress depth and camera pose. However, we start from 3D voxel center and end up at 2D pixel to only regress the depth model while taking camera pose as prior, which is why SFM-based self-supervised depth estimation has scale ambiguity, while our SDF estimation is directly in real scale. 

\noindent
\textbf{SDF Co-Planar Loss}. Photometric constraints are insufficient for indoors scenes due to large non-textured regions and in-plane rotations. Thus, we take advantage of the special geometric constraints in indoor scenes. Most indoor scenes have large planes such as walls, floors, and ceilings, where textures within such planes are mostly similar. Inspired by \cite{p2net, planercnn, planenet, piece-wise} that implement planar constraints in 2D depth maps, we extend it to 3D SDF. Specifically, we adopted `Felzenszwalb superpixel segmentation' \cite{plane_seg} to extract `super-pixels', which covers piece-wise large group of regions that have low pixel intensity gradients, which are considered as a planar region. The algorithm uses greedy search to extract super-pixels and is free of learning. Based on the planar segmentation and the depth planar constraints from \cite{structdepth}, we propose a voxel-SDF driven co-planar loss.

Our goal is to derive plane parameters under planar constraints, and learn the plane parameters in a self-supervised manner. Specifically, the plane segmentation extracts $n$ super-pixels from a 2D image, with each super-pixel corresponding to a continuous plane. For the 2D projected voxel center point P (as shown in Figure \ref{fig:sdf photometric loss}), if it belongs to super-pixel $SP_m$ in the 2D plane, then the surface 3D point $S$ corresponding to $P$ also belongs to the surface plane of class $m$ in 3D space. Using the surface point $S$, the plane $m$ can be defined as Eq. \ref{eq:plane_onepoint}, where $\hat{s}_{0}$ is an estimated surface point in the plane, and $A_m$ is the plane parameter.

\vspace{-3mm}
\begin{equation}
    A_{m}^T \hat{s}_{0} = 1
    \label{eq:plane_onepoint}
\end{equation}
\vspace{-5mm}

While using only one 3D point to simulate a plane is ill-posed, a large number of estimated 3D surface points are obtained by projecting voxel centers to different camera views. With $n$ 2D projected points $p_1$, $p_2$ ...... $p_n$ belonging to super-pixel $SP_m$, there are $n$ 3D surface points $s_1$, $s_2$, ......, $s_n$ belonging to 3D surface plane $m$. Eq. \ref{eq:plane_onepoint} is extended to Eq. \ref{eq:plane_npoint}, where $\hat{S}_{n} = [\hat{s}_{1}, \hat{s}_{2}, ......, \hat{s}_{n}]$, and $Y_m = \Vec{1} = [1,1,...,1]$.

\vspace{-3mm}
\begin{equation}
    \hat{S}_{n} A_{m}^T  = Y_{m}
    \label{eq:plane_npoint}
\end{equation}
\vspace{-5mm}

The plane parameter $A_{m}$ is then estimated by least-square method as Eq. \ref{eq:least_square}, where $\epsilon$ is a small scalar for stability, and $I$ is an identity matrix.

\vspace{-3mm}
\begin{equation}
    A_m = (\hat{S}_n^T\hat{S}_n+\epsilon I)^{-1}\hat{S}_n^TY_m
    \label{eq:least_square}
\end{equation}
\vspace{-5mm}

With the estimated plane parameter, the pseudo surface points can be retrieve  by $\hat{S_n}' = (A_m^T \hat{S_n})^{-1}$. The pseudo surface and estimated surface are expected to align together, and we implement such a co-planar geometric constraint as the self-supervised co-planar SDF loss as Eq. \ref{eq:plane_loss}.

\vspace{-3mm}
\begin{equation}
    L_{plane} = \sum_{M}\sum_{N}{|\hat{S_n}-\hat{S_n}'|}
\label{eq:plane_loss}
\end{equation}
\vspace{-3mm}

\noindent
\textbf{Depth Consistency Loss.} We also propose depth consistency loss to further boost SDF from NeRF. Specifically, we estimate sparse Pseudo-SDF depth for target views from estimated SDF (as Figure \ref{fig:sdf photometric loss}), and render NeRF-depth for corresponding target views. Since Pseudo-SDF depth is in real scale, we first use it to recover NeRF-depth's scale, and enforce consistency between the two estimated depths. 

\vspace{-3mm}
\begin{equation}
    L_{depth} = \sum_{N}\sum_{D \in cam}{|\hat{D_{sdf}}-\hat{D_{NeRF}}|}
\label{eq:depth_loss}
\end{equation}
\vspace{-3mm}

\noindent
where $\hat{D_{sdf}}$ and $\hat{D_{NeRF}}$ are Pseudo-SDF depth and the scale-recoverd NeRF-depth, respectively.

\noindent
\textbf{Total Loss.} We implement standard NeRF losses for NeRF encoder, including RGB consistency with input images, SSIM, and smooth loss, and we jointly train everything end-to-end in pure self-supervision.

\vspace{-8mm}
\begin{equation}
    L_{NeRF} = L_{rgb} + L_{smooth} + (1 - SSIM)
\label{eq:nerf_loss}
\end{equation}
\vspace{-6mm}

\noindent
The total loss is the weighted sum of all losses, where $\lambda$s are the weights,

\vspace{-3mm}
\begin{multline}
    L_{total} = \lambda_{sdf}L_{sdf} + \lambda_{plane}L_{plane} \\ + 
    \lambda_{depth}L_{depth} + \lambda_{NeRF}L_{NeRF}
\label{eq:total_loss}
\end{multline}
\vspace{-5mm}
\section{Experiments}
%\subsection{Datasets \& Implementation Details}
\noindent
\textbf{Datasets.}
We conduct our experiments on ScanNetV2 \cite{scannet} and 7Scenes datasets \cite{7scenes}  for training and evaluation. ScanNet is a real dataset for indoor 3DR, which consists of 2.5M images captured in 1513 different indoor scenes. 7Scenes is another challenging indoor real dataset, consisting of  7 scenes, with rich contents of RGB, depth, and camera pose.

\noindent
\textbf{Metrics.}
We use the same evaluation metrics as Atlas \cite{atlas} and NeuralRecon \cite{neucon}, and we evaluate both 3D geometric mesh metrics and 2D rendered depth metrics. As in \cite{atlas}, since there is no depth directly estimated from the model, we project 3D mesh to get 2D depth maps and use the rendered depth maps for 2D metrics evaluation.

\noindent
\textbf{Implementation Details.}
The weights for each loss ``$\lambda_{sdf}$, $\lambda_{plane}$, $\lambda_{depth}$, $\lambda_{NeRF}$'' are ``1, 0.05, 1, 1'', respectively. For fair comparison with previous voxel-SDF works, we use the same voxel configurations where voxel size is set to $4cm$, scene fragment resolution is set to $96\times96\times96$, SDF truncation distance is set to $30cm$, max depth for ground truth SDF fusion is set to $3m$. We train and test on all samples based on standard training/testing splits of ScanNet where entire framework is trained end-to-end in pure self-supervision from scratch with randomly initialized weights, except for the image encoder Resnet-18, which we choose to be light and efficient. We train on randomly separated scene fragments over all scenes for the first 20 epochs without GRU, which is to reinforce the inner-consistency within fragments and warmup for each part of the framework. We then train on each continuous scene fragments over all scenes together with GRU to reinforce the inter-consistency between fragment. We use 7Scenes only for few-shot transfer and testing. We use NeRF decoder only in training to boost SDF estimation, and we only keep SDF for testing and inference. 

\subsection{Evaluation Results}
\noindent
\textbf{Baselines.} We conduct evaluation and comparison with monocular 3DR in Table \ref{table:3dr class}. Although non-generalizable methods\cite{neuralangelo,neuralwarp,neuris} generate accurate results, they train with one scene and can be only tested on the same scene, which is under completely different training/testing setup to ours, so we do not directly compare with these works. For the generalizable supervised voxel-SDF methods \cite{atlas,neucon}, we directly compare both 3D geometric mesh and 2D rendered depth. Since we get 3D mesh ground truth by fusing SDF with depth map ground truth, to make the comparison rigorous and fair, we get estimated 3D mesh by fusing SDF with estimated depth from self-supervised/supervised depth estimation works. For supervised depth methods \cite{mvdepthnet, gpmvs, dpsnet}, we directly use their depth estimation for SDF fusion, for self-supervised depth \cite{p2net, monoindoor, movingindoor, structdepth}, and since there is scale ambiguity, we first use ground truth depth to recover real scale for estimated depth, and then fuse SDF. NYUV2 is the most popular dataset for indoor depth estimation. Since our SDF regression requires camera pose and NYUV2 does not offer it, we do not test on NYUV2. However, we choose the currently best three self-supervised models \cite{distdepth, structdepth, p2net} from NYUV2 leaderboard for comparison. Although trained on NYUV2, these models claim zero-shot transfer for indoor scenes by testing a subset of ScanNet in their original papers. We test them with the full testing split for comparison with our work. MonoNeRF \cite{mononerf} is our strongest baseline, which also uses the same training/testing splits of ScanNet as in our method. It is the latest self-supervised work to jointly train NeRF with a SFM model, and use NeRF to boost SFM performance.  Our self-supervised voxel-SDF regression is also based on SFM, but MonoNeRF is still depth estimation while our MonoSelfRecon estimates continuous 3D mesh for a whole scene. 

\begin{table}[]
%\resizebox{\textwidth}{!}
\setlength\tabcolsep{1pt}
\footnotesize
\begin{tabular}{llllllll}
\hline
\textbf{2D Depth}         & Train        & AbsRel$\downarrow$ & AbsDiff$\downarrow$ & SqRel$\downarrow$ & RMSE$\downarrow$  & \textbf{$\delta1\uparrow$}  & Comp$\uparrow$  \\ \hline
MVDepthNet\cite{mvdepthnet}     & sup         & 0.098   & 0.191    & 0.061  & 0.293 & 89.6 &  0.928 \\
GPMVS\cite{gpmvs}          & sup          & 0.130   & 0.239    & 0.339  & 0.472 & 90.6 & 0.928 \\
DPSNet\cite{dpsnet}         & sup          & 0.087   & 0.158    & 0.035  & 0.232 & 92.5 & 0.928 \\
COLMAP\cite{colmap}         & sup          & 0.137   & 0.264    & 0.138  & 0.502 & 83.4 & 0.871 \\
Atlas\cite{atlas}          & sup          & 0.065   & 0.123    & 0.045  & 0.251 & 93.6 & 0.999 \\
NeuralRecon\cite{neucon}    & sup          & 0.065   & 0.106    & 0.031  & 0.195 & 94.8 & 0.909 \\
NeuralRecon\cite{neucon}    & weak         & 0.107   & 0.183    & 0.046  & 0.268 & 87.2 & 0.817 \\
\hline
Distdepth\cite{distdepth}   & self      & 0.360   & 0.485    & 0.308  & 0.583 & 48.0 & 0.889 \\
Monodepth2\cite{monodepth2}   & self      & 0.205   & -    & 0.129  & 0.453 & 67.9 & - \\
P2Net\cite{p2net}   & self               & 0.253   & 0.336    & 0.171  & 0.429 & 65.1 & 0.888 \\
Structdepth\cite{structdepth}   & self      & 0.219   & 0.139    & 0.139  & 0.383 & 70.9 & 0.896 \\
MonoNeRF\cite{mononerf}   & self         & 0.169   & -    & 0.089  & 0.375 & 76.0 & - \\
Ours  & self                    &0.143    & 0.231    & 0.090  & 0.333 & 79.2 & 0.872 \\
Ours  & weak                    &0.089    & 0.134    & 0.038  & 0.208 & 89.8 & 0.827 \\\hline
\end{tabular}
\setlength\tabcolsep{3.5pt}
\begin{tabular}{lllllll}
\textbf{3D Mesh}         & Train        & Comp$\downarrow$  & Acc$\downarrow$   & Recall$\uparrow$ & Prec$\uparrow$  & \textbf{F-score$\uparrow$} \\ \hline
MVDepthNet\cite{mvdepthnet}     & sup        & 0.040 & 0.240 & 0.831  & 0.208 & 0.329   \\
GPMVS\cite{gpmvs}          & sup         & 0.031 & 0.879 & 0.871  & 0.188 & 0.304   \\
DPSNet\cite{dpsnet}         & sup         & 0.045 & 0.284 & 0.793  & 0.223 & 0.344   \\
COLMAP\cite{colmap}         & sup         & 0.069 & 0.135 & 0.634  & 0.505 & 0.558   \\
Atlas\cite{atlas}          & sup         & 0.062 & 0.128 & 0.732  & 0.382 & 0.499   \\
NeuralRecon\cite{neucon}    & sup         & 0.120 & 0.062 & 0.428  & 0.592 & 0.494   \\
NeuralRecon\cite{neucon}    & weak         & 0.262 & 0.089 & 0.157  & 0.308 & 0.205   \\
\hline
Distdepth\cite{distdepth}   & self   & 0.262 & 0.371 & 0.159  & 0.104 & 0.121    \\ 
P2Net\cite{p2net}   & self  & 0.170 & 0.264 & 0.231  & 0.148 & 0.179    \\ 
Structdepth\cite{structdepth}   & self      & 0.170 & 0.217 & 0.243  & 0.181 & 0.207    \\ 
Ours   & self                   & 0.212 & 0.185 & 0.262  & 0.263 & 0.260    \\ 
Ours   & weak                   & 0.197 & 0.075 & 0.293  & 0.469 & 0.358    \\ \hline
\end{tabular}
\vspace{-3mm}
\caption{\textbf{2D Depth and 3D Mesh metrics on ScanNet.} ``sup'' ``self'' ``weak'' denote supervised, self-supervised, and weakly supervised training, respectively. We use the same training/testing splits as \cite{atlas, neucon}, where the results of \cite{mvdepthnet,gpmvs,dpsnet,colmap} are directly from \cite{atlas, neucon}'s papers, and we test \cite{distdepth,p2net,structdepth} in this work. Results of \cite{monodepth2, mononerf} are directly from \cite{mononerf} since it also uses same training/testing splits as \cite{atlas,neucon} and our method. Since MonoNeRF\cite{mononerf}'s code is not released, 3D mesh of \cite{mononerf,monodepth2} from Table \ref{table:scannet} are not available, so they are not compared in this table.
}
\label{table:scannet}
\vspace{-7mm}
\end{table}

\noindent
\textbf{ScanNet.} For depth evaluation, we follow the same standards of previous works \cite{atlas, neucon} to render depth from 3D mesh. Table \ref{table:scannet} shows both rendered 2D depth and 3D mesh comparison. Our MonoSelfRecon outperforms all SOTA self-supervised depth estimation from NYUV2 leaderboard, especially the latest strongest baseline MonoNeRF. The 3D mesh comparison shows consistent results to 2D rendered depth. Since MonoNeRF has not released their code, we only compare it for 2D depth and could not generate their 3D mesh for comparison. 

Although our MonoSelfRecon outperforms all SOTA generalizable self-supervised 3DR and is comparable to depth-based purely-supervised 3DR, there is still a clear gap to purely-supervised voxel-SDF methods using fine-grained SDF annotations, so we conduct experiments on weakly-supervised training to further demonstrate our design. While all supervised voxel-SDF methods \cite{atlas,neucon,vortx} use voxel-SDF annotations for a L1 loss, NeuralRecon also uses a coarse binary voxel-mask annotations as weakly supervised signal, where the mask tells if each voxel-SDF is within truncation distance. In our purely self-supervised model setup, we filter out invalid voxels beyond truncation distance at each level of the SDF decoder in both training and testing, and only pass valid voxels to the next level. While for our weakly-supervised training, inspired by NeuralRecon, besides using our self-supervised losses, we add a mask regressor at each level and supervise with coarse mask ground truth, which is only used for training, and the mask regressor estimates valid voxels in testing. As Table \ref{table:scannet} shows, our model improves obviously with coarse mask signal, and is even better than some fully-supervised works. More importantly, we also conduct weakly-supervised training by removing SDF ground truth and only using mask for original NeuralRecon, and we train both works to same epochs. Comparing weakly-supervised results, we clearly outperform NeuralRecon in weak supervision, which validates our purely self-supervised design.

\begin{table}[]
%\small
\footnotesize
\setlength\tabcolsep{2pt}
\begin{tabular}{lllllll}
\hline
\textbf{2D Depth}        & Train    & \textbf{$\delta1\uparrow$}     & AbsRel$\downarrow$ & SqRel$\downarrow$ & RMSE$\downarrow$ & Comp$\uparrow$\\ \hline
DeMoN\cite{demon}   & sup   & 31.9 & 0.389 & 0.420  & 0.855 & -    \\ 
MVSNet\cite{mvsnet}   & sup  & 64.1 & 0.234 & 0.190  & 0.508 & -    \\ 
NeuralRGBD\cite{neuralrgb}   & sup      & 69.3 & 0.176 & 0.112  & 0.440 & -    \\ 
MVDNet\cite{mvdepthnet}   & sup     & 71.8 & 0.193 & 0.235  & 0.459 & -    \\ 
DPSNet\cite{dpsnet}   & sup      & 70.9 & 0.199 & 0.142  & 0.438 & -    \\ 
DeepV2D\cite{deepv2d}     & sup        & 42.8  & 0.437 & 0.553  & 0.869 & -   \\
CNMNet\cite{CNMNet}          & sup         & 76.6 & 0.161 & 0.083  & 0.361 & -   \\
NeuralRecon\cite{neucon}    & sup         & 82.0 & 0.155 & 0.104  & 0.346 & -   \\
Neucon\_weak\cite{neucon}    & weak         & 78.2 & 0.163 & 0.101  & 0.303 & 0.626   \\
Ours(weak)   & weak                   & 68.0 & 0.209 & 0.148  & 0.474 & 0.634    \\ 
Ours(self)   & self                   & 63.3 & 0.219 & 0.155  & 0.519 & 0.818    \\ \hline
\textbf{3D Mesh}         & Train        & \textbf{F-score$\uparrow$}  & Recall$\uparrow$ & Prec$\uparrow$  &  Comp$\downarrow$ & Acc$\downarrow$\\ \hline
DeepV2D\cite{deepv2d}     & sup        & 0.115 & 0.175 & 0.087  & 0.180 & 0.518   \\
CNMNet\cite{CNMNet}          & sup         & 0.149 & 0.246 & 0.111  & 0.150 & 0.398   \\
NeuralRecon\cite{neucon}    & sup         & 0.282 & 0.227 & 0.389  & 0.228 & 0.100   \\
Neucon\_weak\cite{neucon}    & weak         & 0.101 & 0.062 & 0.356  & 0.103 & 0.473   \\
Ours(weak)    & weak         & 0.135 & 0.085 & 0.369  & 0.417 & 0.152   \\
Ours(self)    & self         & 0.146 & 0.101 & 0.274  & 0.323 & 0.215   \\
\hline
\end{tabular}
\vspace{-3mm}
\caption{\textbf{2D depth and 3D mesh metrics on 7Scenes.} ``Ours(weak)'' and ``Ours(self)'' denote ours pre-trained in weak-supervision and self-supervision on ScanNet. ``Neucon\_weak'' denotes NeuralRecon pre-trained in weak-supervision on ScanNet.
}
\label{table:7scenes}
\vspace{-6mm}
\end{table}

\noindent
\textbf{7Scenes.} We also test our models on 7Scenes dataset. To show our model's generalization, as previous works \cite{neucon} we do not train from scratch on 7Scenes, but to do few-shot transfer learning from our pre-trained models on ScanNet. Table \ref{table:7scenes} shows comparisons on both 3D mesh and 2D depth. We conduct three few-shot learnings: 1.Self-supervised transfer from our self-supervised pre-trained model on ScanNet without any annotations, denoted as ``Ours(self)'' in the table; 2.Weakly-supervised transfer from our weakly-supervised pre-trained model on ScanNet with coarse voxel-mask annotations, denoted as ``Ours(weak)'' in the table; 3.Weakly-supervised transfer from the original NeuralRecon pre-trained in weak-supervision on ScanNet with coarse voxel-mask annotations, denoted as ``Neucon\_weak''. Other results in the table are directly from \cite{neucon}. 

The few-shot transfer is done in a few minutes with few frames from 7Scenes training splits. Results show that our purely self-supervised transfer is comparable and even outperforms to fully-supervised methods like DeepV2D \cite{deepv2d} and CNMNet \cite{CNMNet}. More importantly, in 3D mesh results, ``Ours(weak)'' is better than ``Neucon\_weak'' and ``Ours(self)'' is better than ``Ours(weak)'', which indicates that results are better if less coarse mask annotations and more our self-supervised signals are used. The reason is that 7Scenes' scenes are more localized and contain more details than ScanNet. While the voxel-mask annotation successfully reflects coarse representations in ScanNet, it fails to catch the scene details in 7Scenes. This is also proved from 2D depth results that although weakly supervision with mask gets better ``$\delta1$'', the pure self-supervision has the highest ``complete'', which is because the mask annotations are too coarse to guide training and lead to an extremely low recall. These results indicate that our self-supervised losses are more reliable than supervised signals when annotations are coarse, this is significantly useful when the model is generalized to a new domain with limited and coarse annotations. 
\iffalse
We show our detailed per-scene results of 7Scenes datasets in Supplementary. 
\fi

\subsection{Ablation Study}

\vspace{-2mm}
\quad We conduct ablation study to show the importance of each part in our framework, where the results are all trained in self-supervision, and the difference is only in architecture. Table \ref{table:scannet_ablation} shows our ablation study results of 3D mesh and depth on ScanNet. ``naive'' means no GRU/NeRF/Attention used in the framework and is purely trained with proposed self-supervised losses. GRU brings essential improvements to the framework, which shows the importance of consistency between each scene fragment, and not considered in depth-based methods. With MPI-NeRF, our result already outperforms the strongest baseline MonoNeRF \cite{mononerf} which uses similar NeRF as Table \ref{table:scannet} shown, and the Attentional Fusion boosts further improvements. We do not jointly train with NeRF and Attentional Fusion for memory efficiency, and the results is sufficient to show their contributions to the framework.

\subsection{Visual Results}
\vspace{-2mm}

\begin{figure*}
\begin{minipage}{\textwidth}
  \centerline{\includegraphics[width=1.0\textwidth]{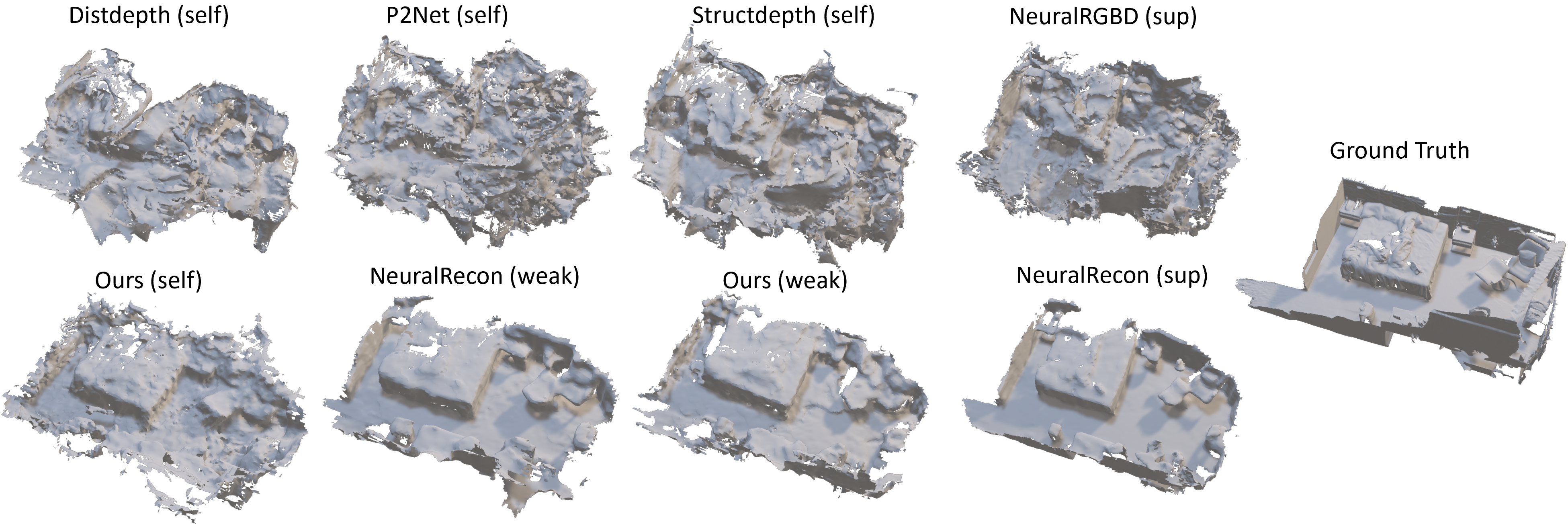}}
\end{minipage}
\vfill
\vspace{-3mm}
\begin{minipage}{\linewidth}
  \centerline{\includegraphics[width=1.0\textwidth]{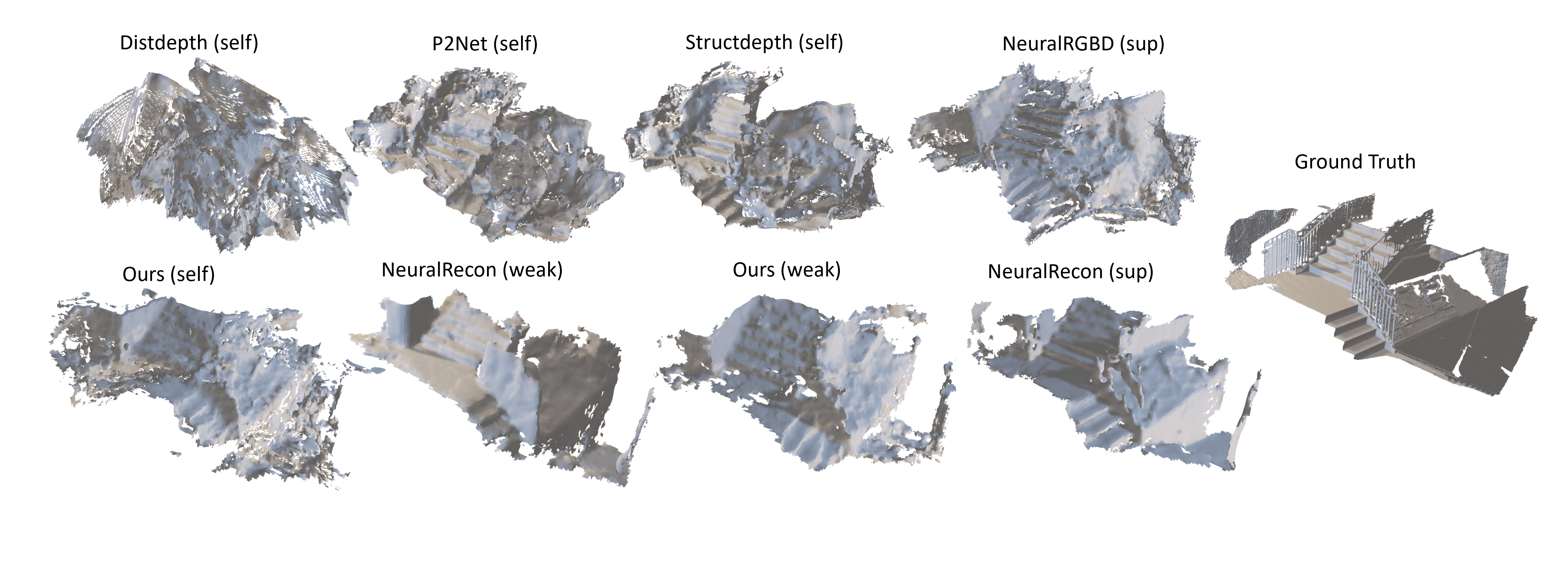}}
\end{minipage}
\vfill
\vspace{-8mm}
\begin{minipage}{\linewidth}
  \centerline{\includegraphics[width=1.0\textwidth]{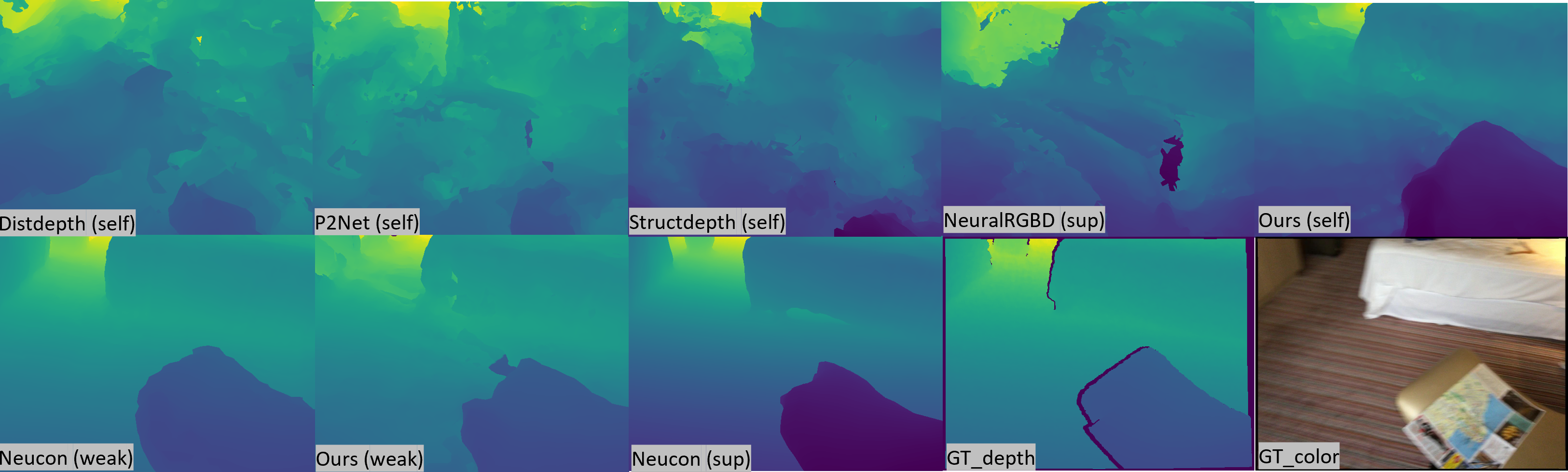}}
\end{minipage}

%\vspace{-3mm}
\begin{minipage}{\linewidth}
  \centerline{\includegraphics[width=1.0\textwidth]{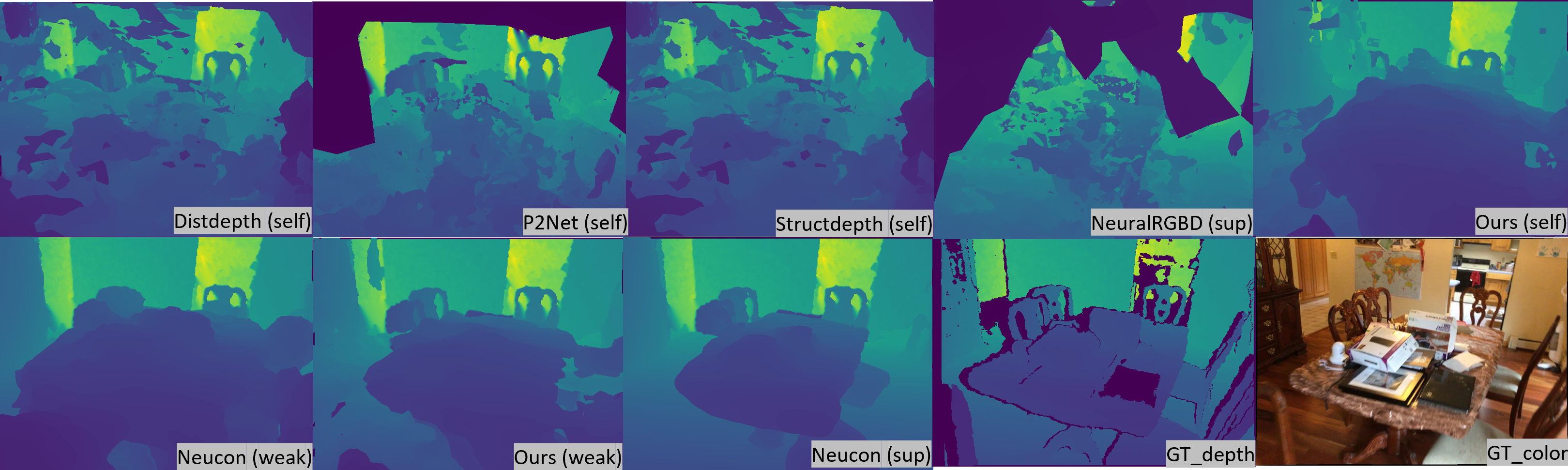}}
\end{minipage}
\vspace{-3mm}
\caption{\textbf{Visual Results on ScanNet.} 3D meshes are shown at top, 2D rendered depth maps are shown at bottom. Our self-supervised results are clearly better than SOTA self-supervised methods on challenging cases and fuzzy RGB input, even better than supervised depth estimation for a few cases. With weak supervision, our result is clearly better than NeuralRecon with weak supervision, which demonstrates our self-supervised design.}
\label{fig:visual results}
\end{figure*}

\quad Figure \ref{fig:visual results} shows 3D mesh (upper part) and rendered 2D depth (lower part) visual comparisons, which are compatible with numerical results. Clearly, our result outperforms SOTA indoor self-supervised depth estimation including Distdepth \cite{distdepth}, P2Net \cite{p2net} and Structdepth \cite{structdepth}, and is even better than supervised work NeuralRGBD \cite{neuralrgb}. For purely self-supervision, only Structdepth can reconstruct some details, but ours can show most of the scene/object details, like the bed, chairs and nightstand in the 1st mesh scene, and the stairs in the 2nd mesh scene. With coarse mask ground truth for weakly supervision, our mesh surface becomes much smoother and does not have much difference to the strongest supervised NeuralRecon trained with fine-grained SDF annotations. The 2D rendered depths at bottom are more straightforward to show details. In the 1st scene, our self-supervised estimation clearly shows the golden chair and the bed, while other self-supervised methods can barely show the contours of the bed. The 2nd scene is more challenging but we are still able to reconstruct fine details such as the farthest chair and the walls at back.

\begin{table}[]
%\resizebox{\textwidth}{!}
\setlength\tabcolsep{1pt}
\footnotesize
\begin{tabular}{lllllll}
\hline
\textbf{2D Depth}    & AbsRel$\downarrow$ & AbsDiff$\downarrow$ & SqRel$\downarrow$ & RMSE$\downarrow$  & $\delta1\uparrow$  & Comp$\uparrow$  \\ \hline
Ours(naive)  &0.358    & 0.678    & 0.417  & 0.846 & 43.4 & 0.817 \\
Ours(+GRU) & 0.176    & 0.277    & 0.127  & 0.387 & 72.8 & 0.786 \\
Ours(+GRU+NeRF)    &0.163    & 0.253    & 0.103  & 0.348 & 76.3 & 0.806 \\
Ours(+GRU+Attention) &0.143    & 0.231    & 0.090  & 0.333 & 79.2 & 0.872 \\ \hline
\end{tabular}
\setlength\tabcolsep{4.5pt}
\begin{tabular}{llllll}
\textbf{3D Mesh}     & Comp$\downarrow$  & Acc$\downarrow$   & Recall$\uparrow$ & Prec$\uparrow$  & F-score$\uparrow$ \\ \hline
Ours(naive)          & 0.278 & 0.206 & 0.198  & 0.155 & 0.171    \\
Ours(GRU)          & 0.212 & 0.226 & 0.247  & 0.197 & 0.217    \\
Ours(GRU+NeRF)          & 0.230 & 0.204 & 0.248  & 0.225 & 0.233    \\
Ours(GRU+Attention)     & 0.212 & 0.185 & 0.262  & 0.263 & 0.260    \\ \hline
\end{tabular}
\vspace{-3mm}
\caption{\textbf{Ablation Study.} ``naive'' denotes no
GRU/NeRF/Attention used in the framework and is purely trained with proposed self-supervised losses.
}
\label{table:scannet_ablation}
\vspace{-7mm}
\end{table}

\section{Conclusion}
\vspace{-2mm}
\quad We propose a novel framework - \textbf{MonoSelfRecon} that for the first time achieves \textbf{explicit 3D mesh} reconstruction for \textbf{generalizable} indoor scenes with monocular RGB views by purely \textbf{self-supervision} on voxel-SDF. We propose novel self-supervised losses corresponding to our novel framework design, and test with thorough experiments on ScanNet and 7Scenes. Our method outperforms the best indoor self-supervised methods so far and is comparable to fully-supervised works. With few-shot learning, it is easily transferred to other domains. The framework is not limited to model design but extensive to any voxel-SDF models, which keeps the advantages of the original model.

{
    \small
    \bibliographystyle{ieeenat_fullname}
    \bibliography{main}
}

\clearpage
\setcounter{page}{1}
\maketitlesupplementary

\section{Relationship with MonoNeRF \cite{mononerf}}
\label{sec:relationship with mononerf}
We discuss the relationship between our strongest baseline MonoNeRF\cite{mononerf} and our proposed method MonoSelfRecon as follows: 1) We share the same idea of SFM-based 3DR with monocular RGB sequence as input, and we both jointly train SFM and a generalizable NeRF, where the NeRF is used to boost SFM performance.  2) Although using SFM as the core of framework design, we regress to different 3D representations, where MonoNeRF regress to view-based 2D depth map while we regress to 3D voxel-based SDF values. 3) While MonoNeRF also jointly estimates camera poses, their 2D view-based depth representation restricts the ability to incrementally complete a whole scene in 3D mesh representation. Fusing TSDF from direct depth estimation is time-consuming, and will cause layered or sparse mesh due to depth inconsistency between each frame. By comparison, our direct voxel-SDF regression enables us to incrementally add the previous mesh to complete the whole scene consistently in mesh representation.  

The mesh representation is a stricter 3D representation over 2D depth map. Theoretically, the depth map can be perfectly rendered from 3D mesh but cannot in reverse, which is further validated by our experiments. Table \ref{table:scannet} and \ref{table:7scenes} show that although all using ground truth for supervised training, the one that directly regresses SDF (NeuralRecon) has a clear advantage on 2D depth metrics over other supervised methods that regresses depth. The reason that although both our method and MonoNeRF are based on SFM while ours outperforms theirs can be also partly attributed to this different 3D representation. Our visual results also reflect this point in Table \ref{fig:visual results}, where although there is no much difference of 2D depth, the difference of 3D mesh is clear. In other words, the depth representation is more visually straightforward than 3D mesh. Consequently, our 3D mesh regressing is a stricter 3D geometric representation than MonoNeRF's 2D depth.

\section{Evaluation Metrics}
We follow the same evaluation metrics as \cite{atlas, neucon}. Details of the metrics are summarized in Table \ref{table:metrics}.
\begin{table}[htb]
\tiny
\setlength\tabcolsep{0.1pt}
\begin{tabular*}{\columnwidth}{@{\extracolsep{\fill}} llll}
\hline
\multicolumn{2}{l}{2D} & \multicolumn{2}{l}{3D} \\ \hline
Abs Rel             & $\frac{1}{n}\sum\left|d-d^*\right|/d^*$      & Acc       & $\text{mean}_{p \in P}\left(\min_{p^* \in P^*}||p-p^*||\right)$\\
Abs Diff            & $\frac{1}{n}\sum\left|d-d^*\right|$          & Comp      & $\text{mean}_{p^* \in P^*}\left(\min_{p \in P}||p-p^*||\right)$\\
Sq Rel              & $\frac{1}{n}\sum\left|d-d^*\right|^2/d^*$    & Prec      & $\text{mean}_{p \in P}\left(\min_{p^* \in P^*}||p-p^*||<.05\right)$\\
RMSE                & $\sqrt{\frac{1}{n}\sum\left|d-d^*\right|^2}$ & Recal     & $\text{mean}_{p^* \in P^*}\left(\min_{p \in P}||p-p^*||<.05\right)$\\
$\tiny\mkern-5mu\sigma<1.25$   & $\frac{1}{n}\sum\left(\max\left(\frac{d}{d^*}, \frac{d^*}{d}\right)<1.25\right)$ & F-score  & $\frac{2 \times \text{Prec} \times \text{Recal}}{\text{Prec} + \text{Recal}}$\\
Comp                & \% valid predictions &                        &                 \\
RMSE log            & $\sqrt{\frac{1}{n}\sum\left|\log(d)-\log(d^*)\right|^2}$   &                        &                 \\
Sc Inv              & $\left(\frac{1}{n}\sum_{i}z_i^2-\frac{1}{n^2}(\sum_i z_i)^2\right)^{1/2}$ &                        &                 \\ \hline
\end{tabular*}
\caption{\textbf{Evaluation Metrics.}}
\label{table:metrics}
\end{table}

\section{Model Details}
\subsection{Attentional View Fusion}
We use a standard Vision Transformer (ViT) Encoder, where we keep the original high-level architecture of the ViT encoder to be: A norm layer, a multi-head attention layer, a norm layer, and a MLP (2 heads are used). Originally the ViT takes image patch/features as input, while we adopted the input to be the nearest 2D features from the projected 3D voxels, which is of size $[N_{view}, N_{points}, C]$, where $N_{view}$ is the number of views in a scene fragment, $N_{points}$ is the number of voxels in a fragment, $C$ is the feature channel. The input also takes the voxel mask as input to filter out the pixels which are invisible to the voxels, and the transformer only takes the visible pixel features. We stack two ViT encoders to update the features, where the output is still of size $[N_{view}, N_{points}, C]$. Then we use a multi-view weighted feature pooling to fuse the updated features at the view channel to 3D features of size $[N_{points}, C]$, where the weight is the number of visible views in a fragment for each voxel. Such design enables more flexibility to adjust the contribution of each view to the 3D voxels. 

\subsection{GRU}
We directly use the GRU module from \cite{neucon}, which is elaborately designed for sparse 3D convolution. The 3D voxel features are obtained by attentional view fusions and fed to the GRU module, where the current 3D fragment features are conditioned on the previous fragment. Using the current 3D global voxel features $G^l_{t}$ and the previous hidden state $H^l_{t-1}$ at layer $l$,  the current hidden state $H^l_{t}$ can be obtained, and the SDF value at each level is regressed from the hidden state $H^l_{t}$. More specifically,

\begin{equation}
\begin{aligned}
     &z_t = \sigma(SparseConv([H^{l}_{t-1}, G^l_t], W_z)) \\
     &r_t = \sigma(SparseConv([H^{l}_{t-1}, G^l_t], W_r)) \\
     &\Tilde{H^l_t} = tanh(SparseConv([r_t \odot H^l_{t-1}, G^l_t], W_h)) \\
     &H^l_t = (1-z_t)\odot H^l_{t-1} + z_{t} \odot \Tilde{H^l_t}
\end{aligned}
\end{equation}

\noindent where $z_t$ is the update gate, $r_t$ is the reset gate, $\sigma$ is the sigmoid function and $W_{*}$ is the weight for sparse convolution.

We first train without GRU within each fragment to warmup the framework with our proposed self-supervised losses, where we call it \textbf{intra-fragment losses}. Because the GRU module is leveraged to enhance the consistency between fragments, the self-supervised learning strategy should be treated differently to intra-fragment losses. There is no need to change the training policy in purely supervised training because SDF ground truth is used, and there is no ground truth in our self-supervision except for the consistency clues between fragments. So we extend the \textbf{inter-fragment losses} to \textbf{intra-fragment losses}. While the model only takes input per fragment, backpropagating whole fragments brings memory challenges, so we only implement the inter-fragment losses on the frames around the boundary of fragments. Specifically, we use the last 4 and first 4 frames of the previous and current fragments to implement the inter-fragment loss.

\begin{figure}
    \includegraphics[width=\linewidth]{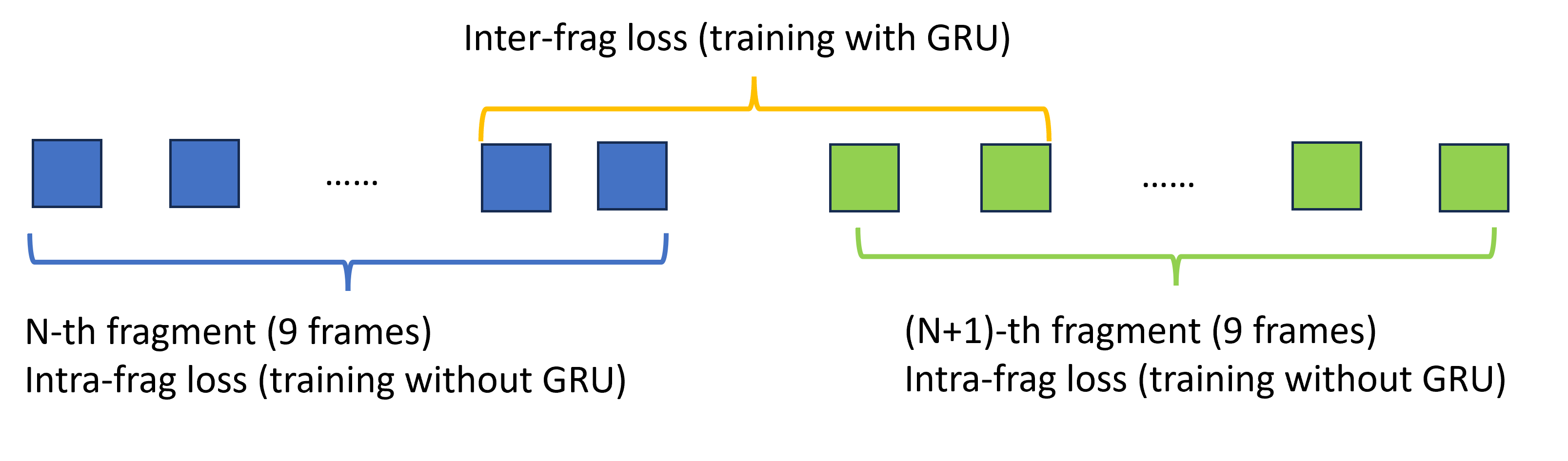}
    \caption{\textbf{Inter/Intra-fragment} losses illustration.}
    \label{fig:gruloss_explain}
\end{figure}

\subsection{NeRF}

\begin{figure}
    \includegraphics[width=\linewidth]{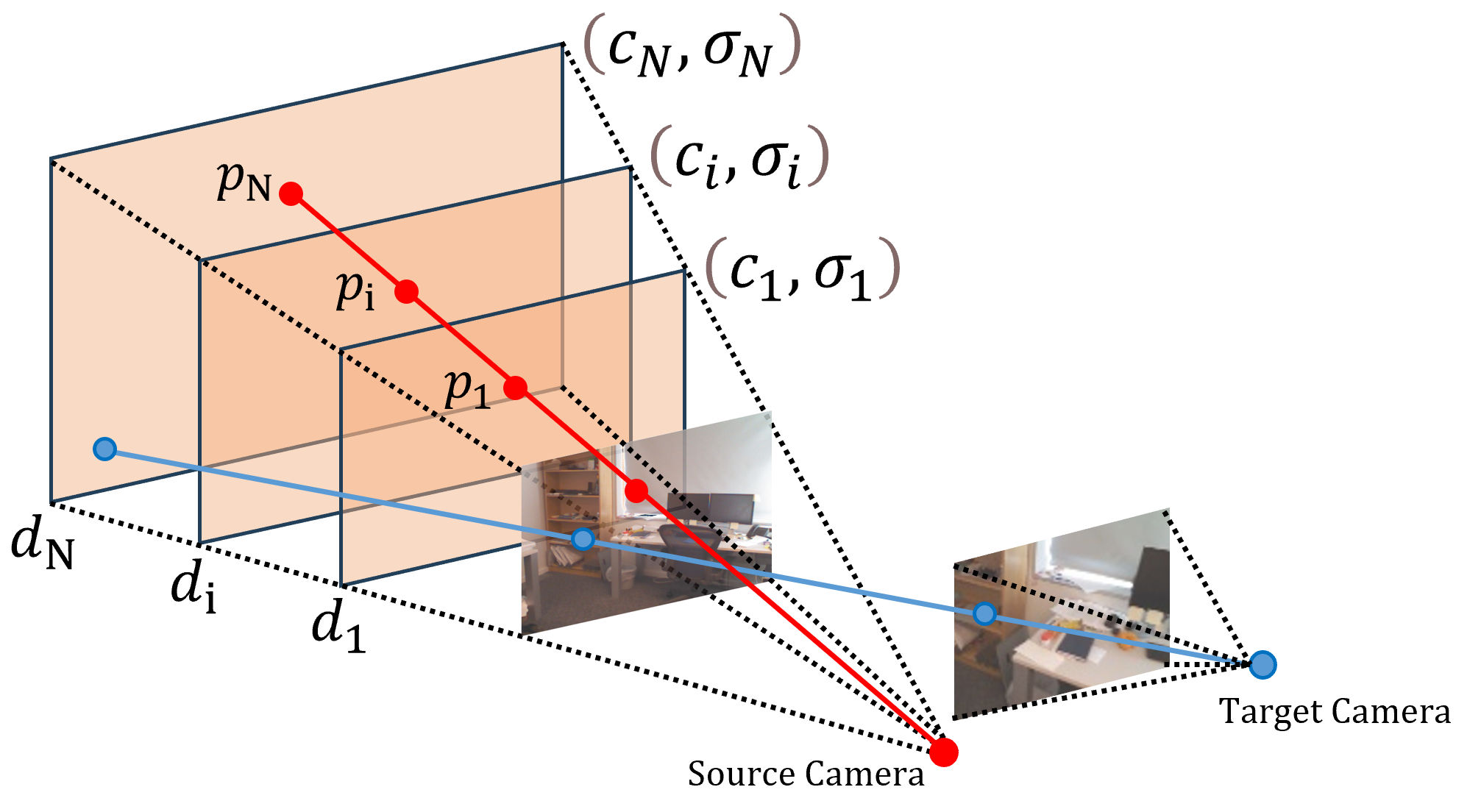}
    \caption{\textbf{Multi-Plane Image (MPI)} NeRF illustration.}
    \label{fig:mpi_explain}
\end{figure}

Since the SDF decoder is generalizable, the NeRF also must be generalizable to boost SDF decoder. For our work, we adopted MPI-NeRF\cite{mpi, mpi-nerf}, which has been directly used by MonoNeRF\cite{mononerf} and proved to be generalizable. As Figure \ref{fig:mpi_explain} shows, in Multi-Plane-Image (MPI) system, an image is represented by a set of parallel planes (orange planes) denoted as RGB-$\sigma$, specifically $(c_{i}, \sigma_{i})_{i=1}^N$,  where the $i_{th}$ plane has $d_{i}$ disparity (reverse of depth) to the camera. The shading points (red points) are selected as the intersection of the parallel planes and the rays shooting from pixels in the image, where $c_{i}$ and $\sigma_{i}$ are the RGB color and density of each shading points at $i_{th}$ plane. In a standard MPI system, the source view RGB image $\hat{I_{s}}$ and depth map $\hat{D_{}}$ can be composed using the ``over'' operation \cite{over_composite} as

\iffalse
\begin{equation}
    \Biggl\{ \begin{array}{cc}
     \hat{I}_{s} = \sum_{i=1}^{D}(c_{i}\sigma_{i}\prod_{j=i+1}^{D}(1-\sigma_{j})) \\
     \\
     \hat{D}_{s} = \sum_{i=1}^{D}(d_{i}^{-1}\sigma_{i}\prod_{j=i+1}^{D}(1-\sigma_{j}))
     \end{array}
\end{equation}
\fi

{
\begin{equation}
\begin{aligned}
    \hat{I}_s &= \sum_{i=1}^{D} (c_i \sigma_i \prod_{j=i+1}^{D} (1 - \sigma_j)) \\
    \hat{D}_s &= \sum_{i=1}^{D} (d_i^{-1} \sigma_i \prod_{j=i+1}^{D} (1 - \sigma_j))
\end{aligned}
\label{eq:combined}
\end{equation}
}

To use MPI system in NeRF style, the composition operation above can be replaced by volumetric rendering \cite{nerf} for both RGB and depth as 
\iffalse
\begin{equation}
    \Biggl\{ \begin{array}{cc}
    \hat{I_s} = \sum_{i=1}^{N}T_{i}(1-exp(-\sigma_{i}\delta_{i}))c_{i} \\
    \\
    \hat{Z_s} = \sum_{i=1}^{N}T_{i}(1-exp(-\sigma_{i}\delta_{i}))z_{i}
    \end{array}
\end{equation}
\fi

{
\begin{equation}
\begin{aligned}
    \hat{I}_s &= \sum_{i=1}^{N}T_{i}(1-exp(-\sigma_{i}\delta_{i}))c_{i} \\
    \hat{Z}_s &= \sum_{i=1}^{N}T_{i}(1-exp(-\sigma_{i}\delta_{i}))z_{i}
\end{aligned}
\label{eq:combined}
\end{equation}
}

\noindent where $z_i$ is the rendered depth (reverse of disparity) $z_i = 1/d_i$, and $\delta_i = ||p_{i+1} - p_i||_2$ is the distance between the two neighbor shading points on a ray. Then we can extend volumetric rendering to target views. First, the corresponding pixels $[u_t, v_t]$ in the target view can be found by 

\begin{equation}
    \begin{bmatrix}
        u_{s} \\
        v_{s} \\
        1
    \end{bmatrix}
    \sim K_{s}(R-tn^{T}d_{i})(K_{t})^{-1}
    \begin{bmatrix}
        u_{t} \\
        v_{t} \\
        1
    \end{bmatrix}
\end{equation}

Here, $[u_s, v_s]$ is the corresponding pixel locations in the source view, $K_s$ and $K_t$ are camera intrinsics of source and target views, $R$ and $t$ are rotation and translation from the target to source view, and $n$ is the norm vector of the $i_{th}$ plane. As the planes are parallel, the RGB $c'_i$ and density $\sigma'_i$ of shading points (blue points) on target rays (blue ray) are equal to those from source rays at the same disparity, as shown in Eq. \ref{mpinerf_samergbsigma},

\iffalse
\begin{equation}
    \Biggl\{ \begin{array}{cc}
    c_{i}'(u_{t}, v_{t}) = c_{i}(u_{s}, v_{s}) \\
    \sigma_{i}'(u_{t}, v_{t}) = \sigma_{i}(u_{s}, v_{s})
    \end{array}
    \label{mpinerf_samergbsigma}
\end{equation}
\fi

{
\begin{equation}
\begin{aligned}
    c_{i}'(u_{t}, v_{t}) = c_{i}(u_{s}, v_{s}) \\
    \sigma_{i}'(u_{t}, v_{t}) = \sigma_{i}(u_{s}, v_{s})
\end{aligned}
\label{mpinerf_samergbsigma}
\end{equation}
}

\iffalse
\begin{equation}
    \Biggl\{ \begin{array}{cc}
     \hat{I}_{t} = \sum_{i=1}^{D}(c_{i}'\sigma'_{i}\prod_{j=i+1}^{D}(1-\sigma_{j}')) \\
     \\
     \hat{D}_{t} = \sum_{i=1}^{D}(d_{i}\sigma'_{i}\prod_{j=i+1}^{D}(1-\sigma_{j}'))
     \end{array}
\end{equation}
\fi

Once we have RGB and density for target views, we can perform volumetric rendering on target views as: 
\iffalse
\begin{equation}
    \Biggl\{ \begin{array}{cc}
    \hat{I_t} = \sum_{i=1}^{N}T_{i}(1-exp(-\sigma'_{i}\delta_{i}))c'_{i} \\
    \\
    \hat{Z_t} = \sum_{i=1}^{N}T_{i}(1-exp(-\sigma'_{i}\delta_{i}))z'_{i}
    \end{array}
\end{equation}
\fi

{
\begin{equation}
\begin{aligned}
    \hat{I_t} = \sum_{i=1}^{N}T_{i}(1-exp(-\sigma'_{i}\delta_{i}))c'_{i} \\
    \hat{Z_t} = \sum_{i=1}^{N}T_{i}(1-exp(-\sigma'_{i}\delta_{i}))z'_{i}
\end{aligned}
\label{eq:combined}
\end{equation}
}

We use standard NeRF RGB loss, where $\hat{I}_s$ and $\hat{I}_t$ are self-supervised with their corresponding input images with a L1 loss. Since we directly use the reverse of disparity for depth, the depth value is scale-ambiguous. As mentioned in the paper, since there is no depth ground truth for pure self-supervision, we use SDF-depth as pseudo-depth to first recover the real scale of $\hat{Z}_s$ and $\hat{Z}_t$, then we impose a consistency loss between $\hat{Z}$ and SDF-depth to boost SDF decoder.

\section{Visual Results}
\begin{figure*}
\begin{minipage}{\textwidth}
  \centerline{\includegraphics[width=1.0\textwidth]{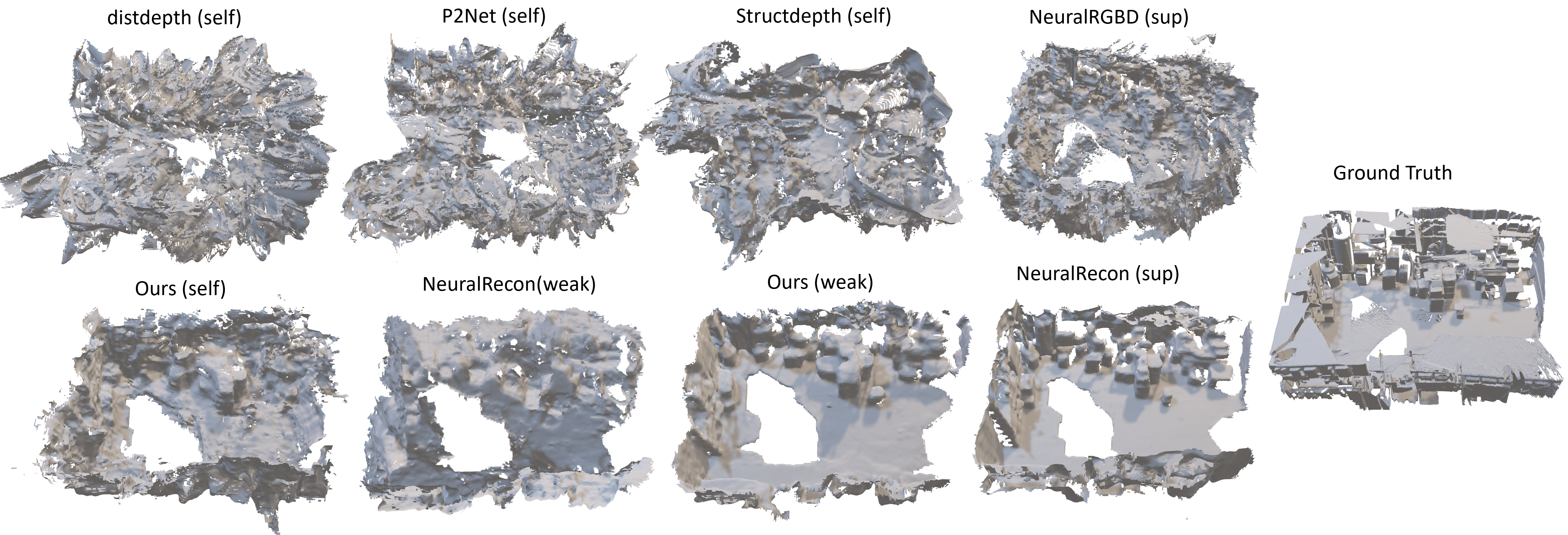}}
\end{minipage}
\vfill
\begin{minipage}{\linewidth}
  \centerline{\includegraphics[width=1.0\textwidth]{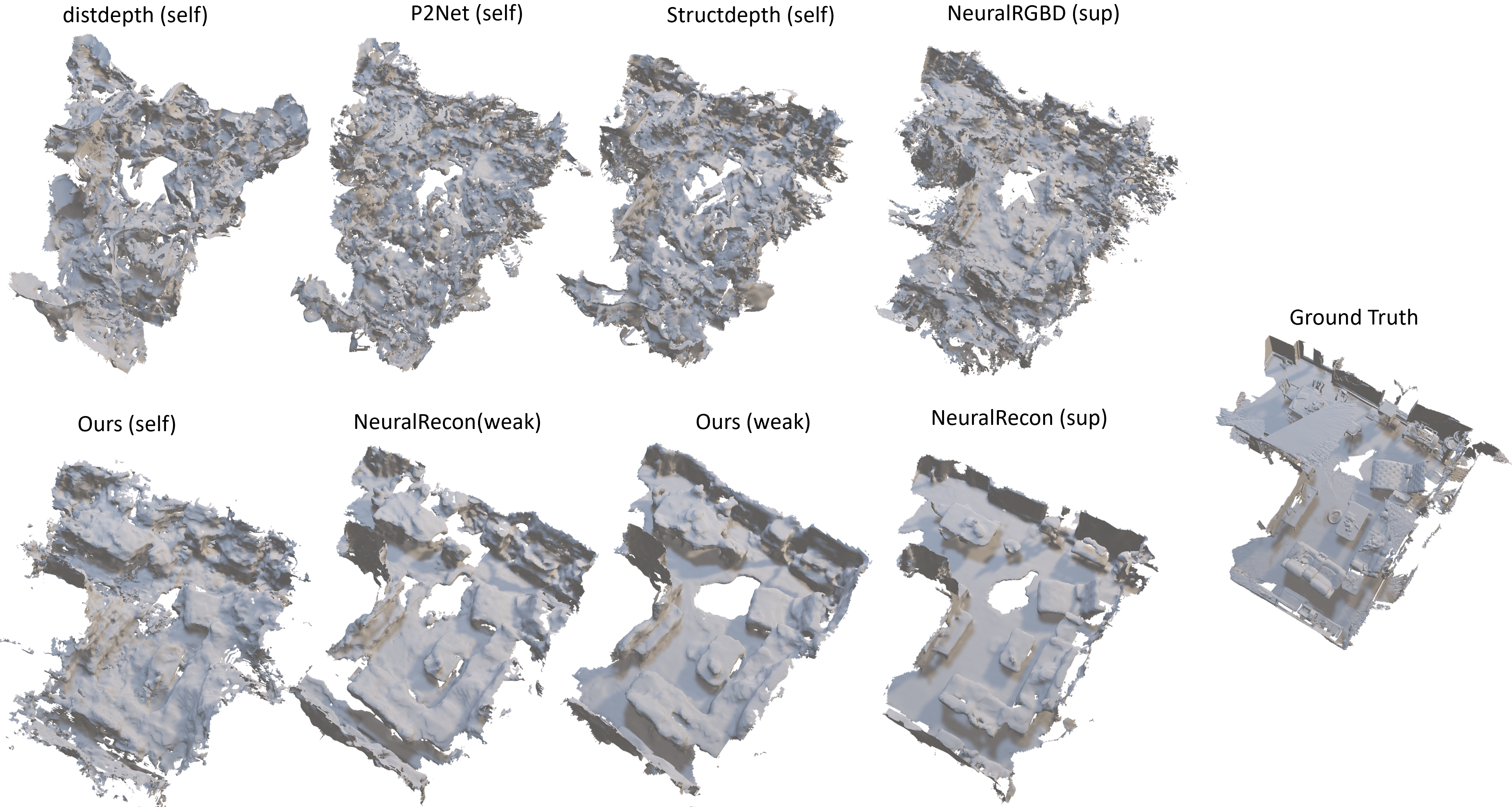}}
\end{minipage}

\iffalse
\vfill
\begin{minipage}{\linewidth}
  \centerline{\includegraphics[width=1.0\textwidth]{figures/scannet_mesh/710.png}}
\end{minipage}
\fi
\label{fig:scannet_supp}
    \caption{Visual Results}
\end{figure*}

We show more visual results of 2D rendered depth and 3D mesh in Figure \ref{fig:scannet_supp}. \textbf{We also attach a PowerPoint file with visual results, where reviewers can rotate and zoom the 3D mesh to see the details},

\section{Limitation}
Although our work combines the advantages of ``self-supervised'' ``generalizable'' and ``3D explicit mesh'' altogether, there are still limitations. So far our MonoSelfRecon can be only used for indoor environments, because we pre-define the 3D scene fragment with a fixed voxel number. Unlike indoor 2D images where depth vary within few meters, the depth can vary significantly just within a single 2D image in outdoor. It is applicable to keep the voxel number while increasing the voxel size, but it will lead to very poor resolution within voxels, which misses most of the details. Moreover, since we regress SDF corresponding to the discrete $N\times N\times N$ voxels of scene fragment, we cannot directly estimate SDF of a continuous 3D space, unless by interpolation. By contrast, SDF-NeRF based methods estimate SDF values in continuous 3D space but it is not generalizable to another scene. Our future works will explore to make SDF-NeRF generalizable, so that the 3DR can be ``self-supervised", ``generalizable", ``explicit'', ``indoor/outdoor'', and ``continous in 3D space''. 

\newpage
\begin{figure*}
\begin{minipage}{\textwidth}
  \centerline{\includegraphics[width=1.0\textwidth]{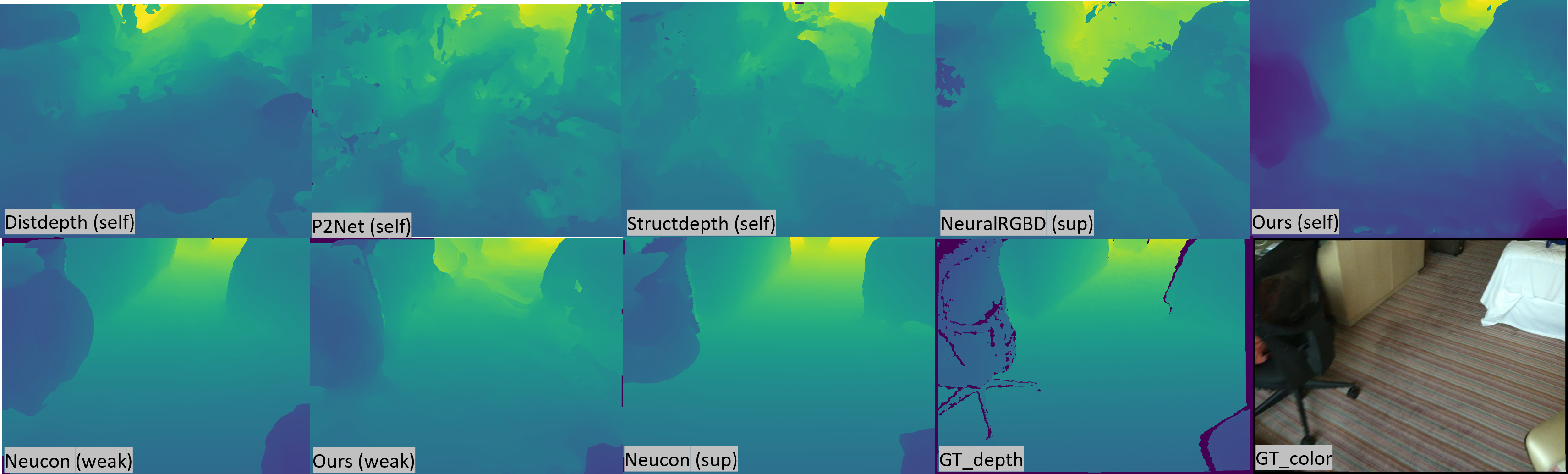}}
\end{minipage}
\vfill
\begin{minipage}{\linewidth}
  \centerline{\includegraphics[width=1.0\textwidth]{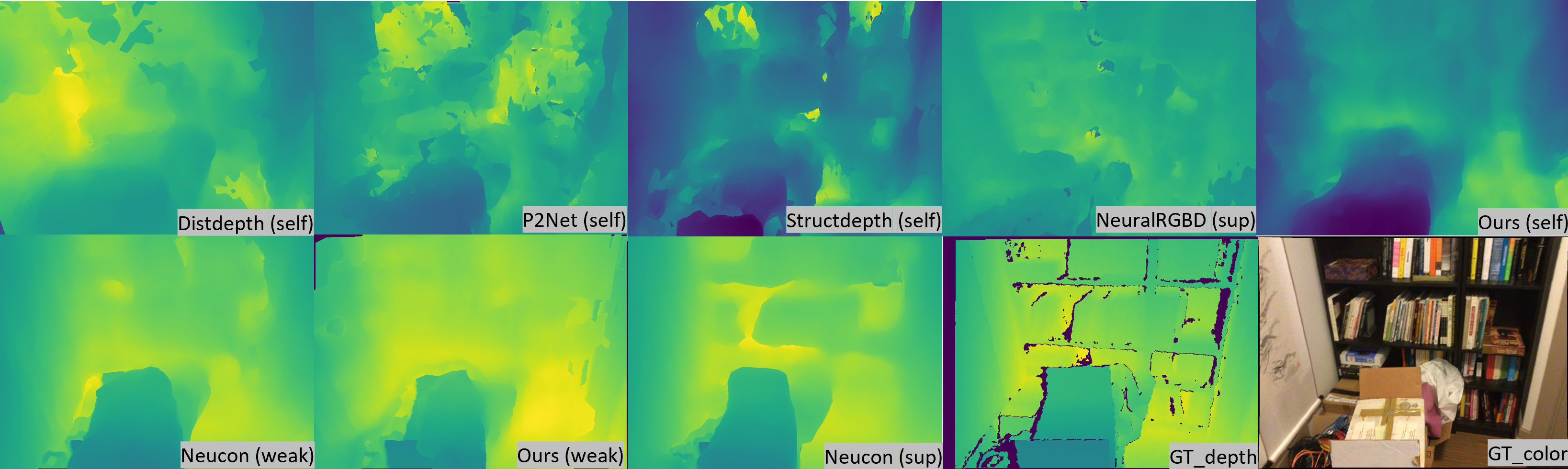}}
\end{minipage}
\vfill
\begin{minipage}{\linewidth}
  \centerline{\includegraphics[width=1.0\textwidth]{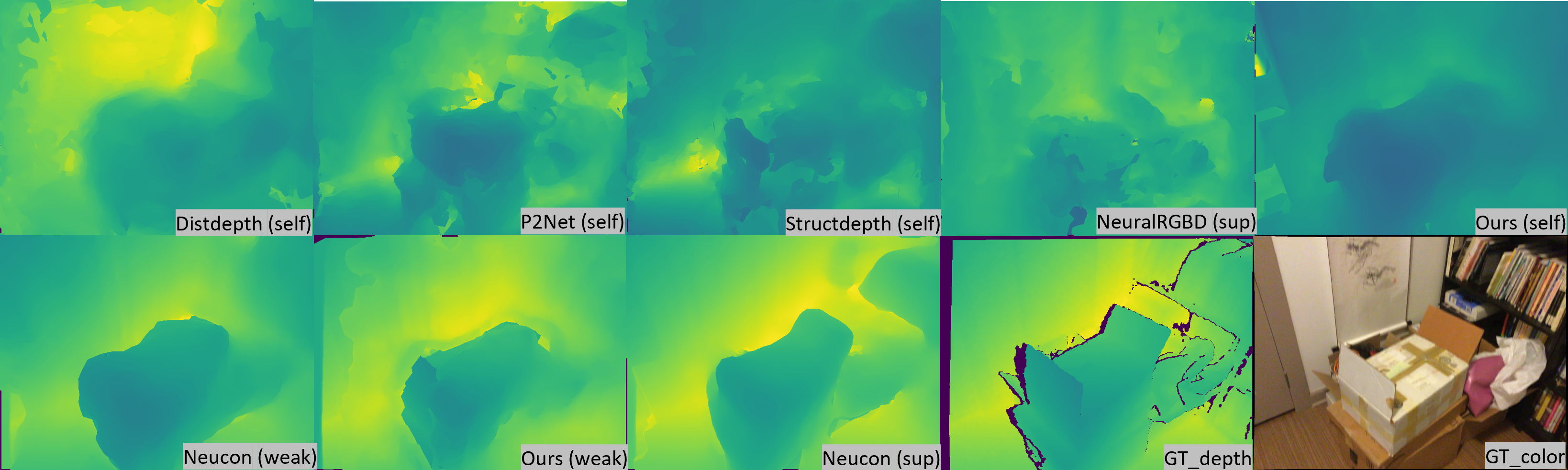}}
\end{minipage}
%\vspace{-3mm}
\begin{minipage}{\linewidth}
  \centerline{\includegraphics[width=1.0\textwidth]{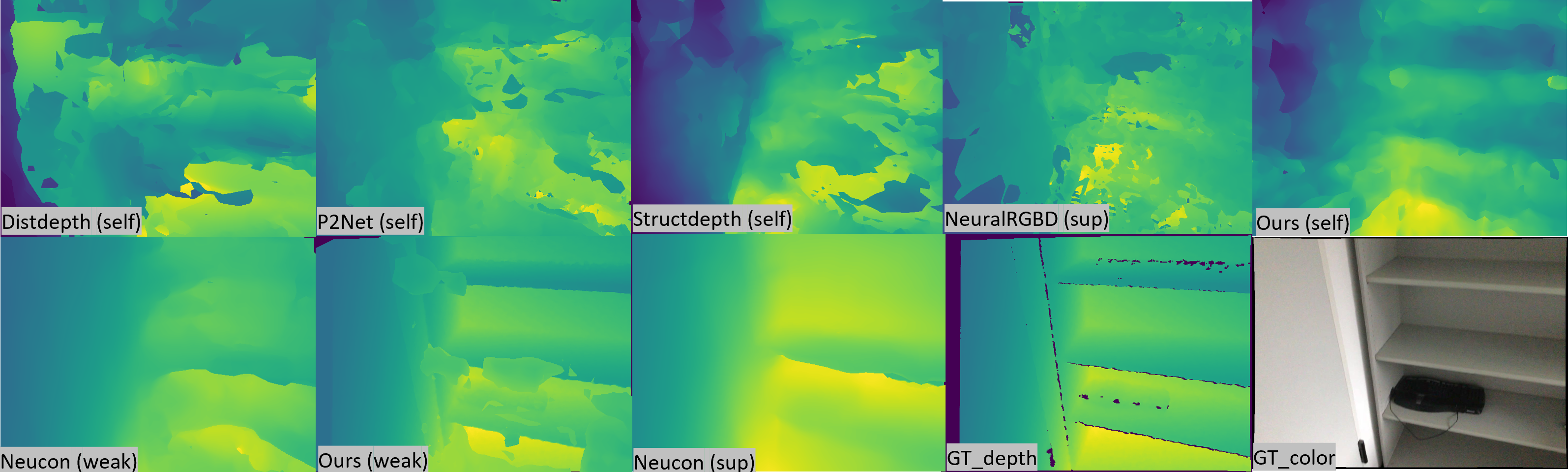}}
\end{minipage}
%\vspace{-3mm}
\end{figure*}

\newpage
\begin{figure*}
\begin{minipage}{\textwidth}
  \centerline{\includegraphics[width=1.0\textwidth]{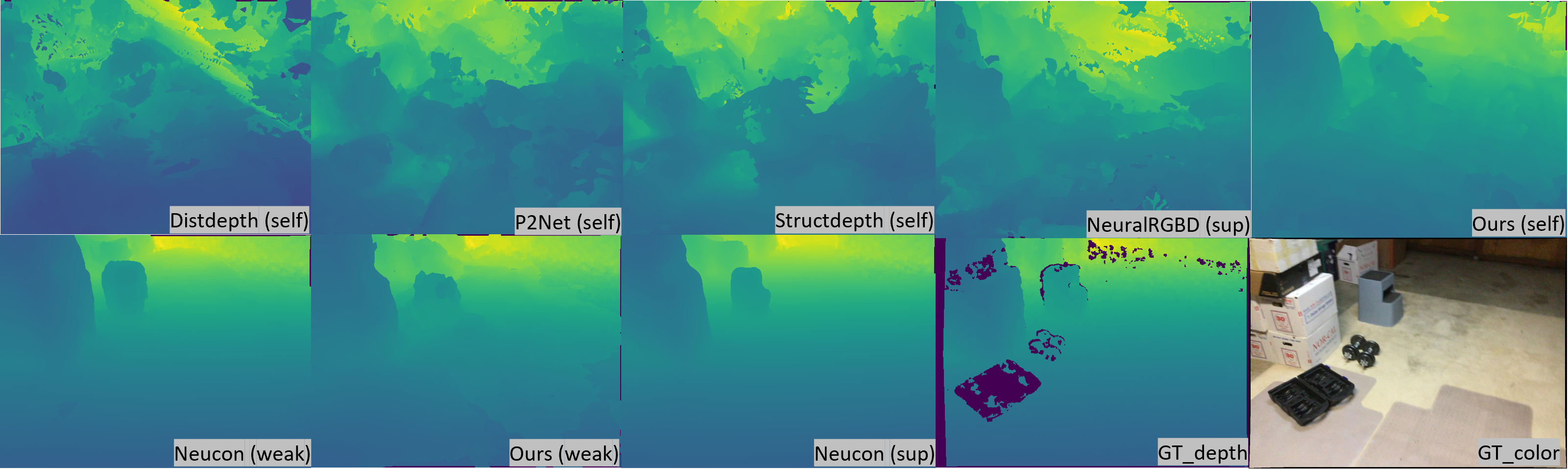}}
\end{minipage}
\vfill
\begin{minipage}{\linewidth}
  \centerline{\includegraphics[width=1.0\textwidth]{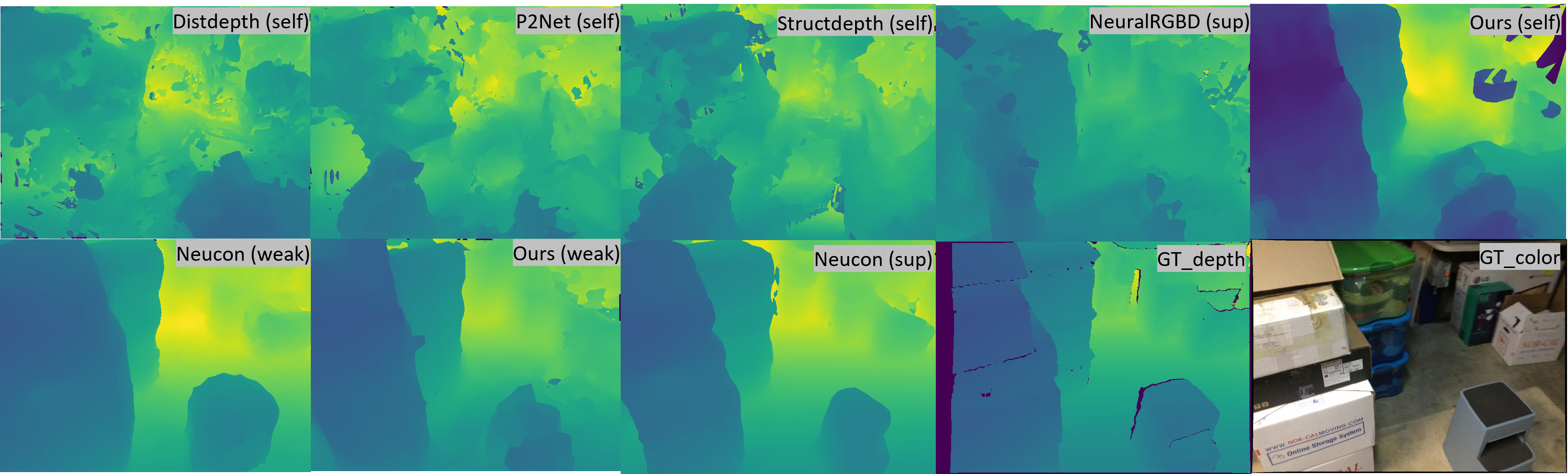}}
\end{minipage}
\vfill
\begin{minipage}{\linewidth}
  \centerline{\includegraphics[width=1.0\textwidth]{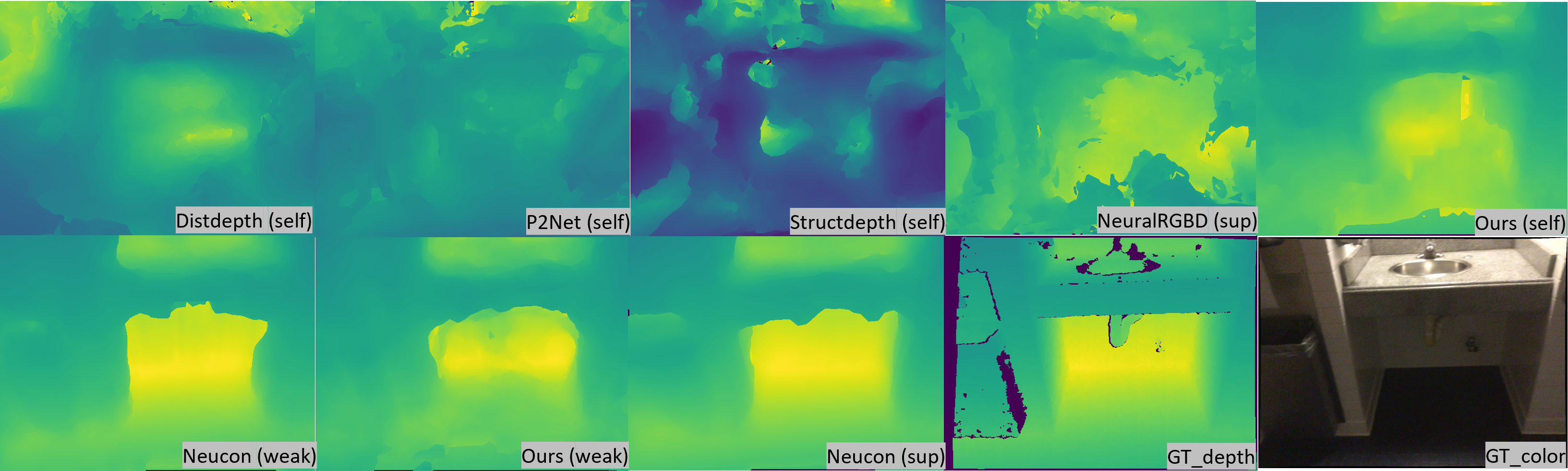}}
\end{minipage}
%\vspace{-3mm}
\begin{minipage}{\linewidth}
  \centerline{\includegraphics[width=1.0\textwidth]{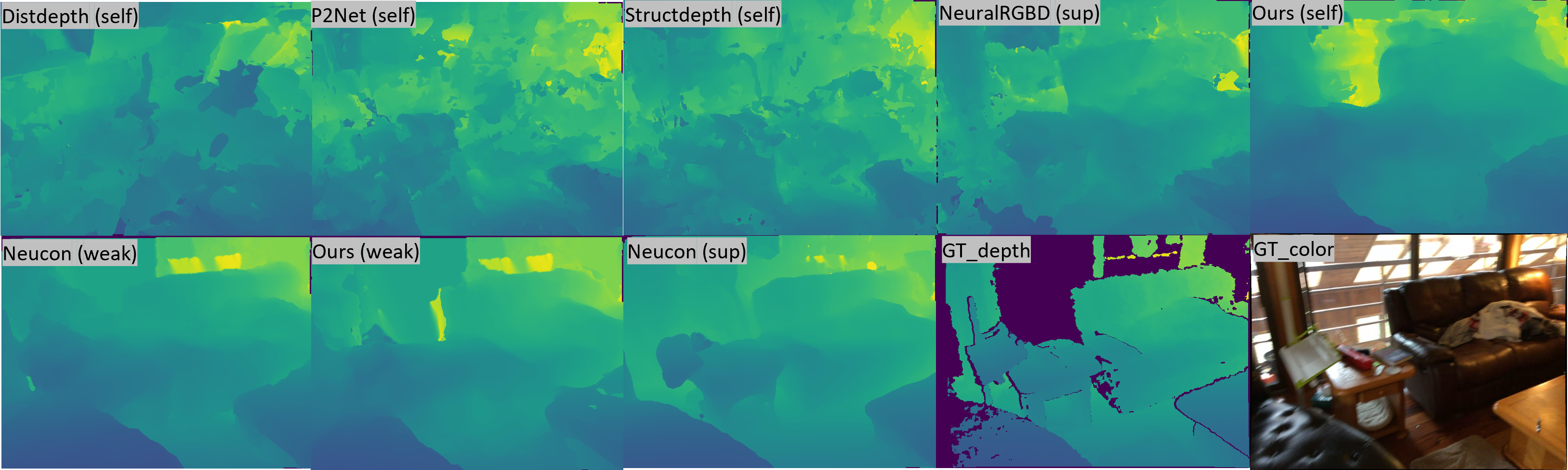}}
\end{minipage}
%\vspace{-3mm}
\vspace{-3mm}
\end{figure*}

\newpage

\end{document}